\documentclass[11pt]{article}

\usepackage[preprint]{acl}

\usepackage{times}
\usepackage{latexsym}

\usepackage[T1]{fontenc}

\usepackage[utf8]{inputenc}

\usepackage{microtype}

\usepackage{inconsolata}

\usepackage{graphicx}

\usepackage{amsmath}  
\usepackage{pifont}
\usepackage{bm}       
\usepackage{bbm}  
\usepackage{booktabs}  
\usepackage{makecell} 
\usepackage{enumitem}
\usepackage{listings}
\usepackage{float}   

\usepackage{soul} 
\usepackage{color} 
\usepackage[svgnames]{xcolor} 

\usepackage{todonotes} 

\newcommand{\plotdesc}{Top: log-odds ratios and confidence intervals obtained using the GLMM; dashed vertical line marking the null-effect (i.e. OR = 1). Bottom left: variant performance deltas ($\Delta_{var}$) in percentage points (pp); hatching marks non-significance. Bottom right: P~values (prior to Holm-Bonferroni correction); dashed vertical line marking the standard statistical significance threshold $\alpha=0.05$.}
\newcommand{\pbdesc}{P~values falling below the standard significance threshold $\alpha=0.05$ marked in bold.}
\newcommand{\ddagdesc}{$\ddagger$ marks cases where Wald-derived statistical significance does not match the bootstrap result.}

\title{The Importance of Being Statistically Earnest:\\A Critical Re-evaluation of GSM-Symbolic}

\author{
  \textbf{Dominika Agnieszka D{\l}ugosz}\textsuperscript{1,2}, 
  \textbf{Arlindo L. Oliveira}\textsuperscript{1},  
  \textbf{Natalia D\'{i}az-Rodr\'{i}guez}\textsuperscript{2}
\\
\\
  \textsuperscript{1}Instituto Superior T\'{e}cnico \& INESC-ID, Universidade de Lisboa, Portugal\\
  \textsuperscript{2}Dept. of Computer Science and AI \& DaSCI Institute, Universidad de Granada, Spain\\
\\
  \small{\textbf{Correspondence:} \href{mailto:dominika.dlugosz@tecnico.ulisboa.pt}{dominika.dlugosz@tecnico.ulisboa.pt}}
}

\begin{document}
\maketitle
\begin{abstract}
The GSM-Symbolic benchmark \citep{mirzadeh2025gsm} reported consistent performance drops across 25 Large Language Models (LLMs) when tested on template-generated variants of GSM8K problems, concluding that the models lack genuine reasoning capabilities. We argue that this conclusion rests on shaky statistical ground. Re-evaluating 20 open-weight models using bootstrapped Generalised Linear Mixed Models with per-question random effects, we find that only 8 exhibit statistically significant performance changes under the original prompt format. Moreover, we identify a previously unacknowledged factor: the distribution of integers in problem texts of the main GSM-Symbolic dataset is systematically shifted towards larger values relative to the original GSM8K (K-S statistic = 0.12, p < 0.001), contradicting the original authors' claims. Controlling for this \textit{large-number effect} accounts for significance in half of the remaining cases. Among models with statistically significant performance deltas, we identify distinct, model-specific behavioural failure profiles --- including fragility of variable binding, arithmetic limitations, and dual-task interference --- underscoring that blanket claims about LLM reasoning risk being both statistically premature and mechanistically misleading.

\end{abstract}

\section{Introduction}

As Large Language Models (LLMs) are developed towards increased robustness in reasoning, reliable evaluation of their problem-solving capabilities is becoming instrumental. Several benchmarking datasets have been proposed to facilitate principled assessment of LLMs, including GSM8K --- a dataset of 8.5k mathematical problems at grade school level \citep{cobbe2021gsm8k}.

However, as new models are released, they might be exposed to the publicly available benchmarking datasets at training time, causing data contamination and possibly inflating the reported performance. Evaluating data contamination effects was the motivation behind the GSM-Symbolic study \citep{mirzadeh2025gsm}. The researchers extracted 100 questions from the GSM8K benchmark and, using carefully constructed templates to vary names and numbers involved, prepared 50 new variants of each of these questions, creating a dataset of 5000 examples\footnote{\citeauthor{mirzadeh2025gsm} additionally studied question variants with increased difficulties and added irrelevant clauses, which are out of the scope of this paper.}. Having tested 25 models of varying architectures and sizes on the dataset, the authors observed a consistent performance decline on the new variants compared to the original 100 questions and hypothesised that this result demonstrates limitations in logical reasoning capabilities of LLMs.

In this work, we revisit the GSM-Symbolic study to quantitatively assess the extent of the reported deterioration and provide more insights concerning possible failure mechanisms.
Our re-evaluation reveals that the per-model accuracy deltas often fail to reach statistical significance ($p < 0.05$) when subjected to rigorous hypothesis testing.
Furthermore, focusing on the models which do exhibit significant performance changes, we propose alternative prompt formats to help isolate potential contributing factors: reliance on the exact strings of words from the GSM8K questions (as assumed by \citeauthor{mirzadeh2025gsm}), difficulties in object attribute tracking, and limitations of arithmetic nature.

\section{Related Work}
Despite long-standing calls for rigorous hypothesis testing in Natural Language Processing (NLP) evaluation \citep{marie2021scientific,dror-etal-2018-hitchhikers}, the absence of statistical significance testing remains a persistent methodological concern. 
\citet{Arias_Blocher_Rodemann_Assenmacher_Jansen_2025} observed a declining trend in the use of statistical validation of evaluations of LLM-generated text over the 2022--2024 period.
\citet{vaugrante2024loomingreplicationcrisisevaluating}'s reevaluation of several advanced prompting techniques revealed a general lack of statistically significant differences when compared to the respective baselines. The authors suggest that the field might be approaching a \textit{replication crisis}, mirroring similar historical occurrences in psychology \citep{Simmons_Nelson_Simonsohn_2011, estimating2015} and medicine \citep{ioannidis2005most, Begley_Ellis_2012}.

First described by \citet{clark1973language}, the \textit{language-as-fixed-effect fallacy} is a type of statistical error known in psychological and linguistic research, referring to treating specific language materials as fixed rather than random effects. By ignoring the inherent, idiosyncratic variance between individual linguistic items (e.g. particular words or sentences), researchers risk making broad claims about language processing that are actually false positives driven by the specific sample used.
To mitigate this fallacy, the field has shifted to mixed-effect modelling of the experimental results \citep{baayen2008mixed}, e.g. through Generalised Linear Mixed Models (GLMMs) \citep{jaeger2008categorical}. GLMMs extend generalised linear models (e.g. logistic regression) by incorporating both \textit{fixed effects} --- population-level factors of theoretical interest --- and \textit{random effects} that capture variability across items or subjects, enabling statistically valid inference over heterogeneous, non-independent data.
More recently, \citet{van2019best} and \citet{card2020little} argued for adopting mixed models in the fields of Natural Language Processing and Generation. 

Reliable evaluation of logical and mathematical reasoning capabilities of LLMs has been the motivation behind multiple benchmarks, such as MATH \cite{hendrycks2021measuring}, BIG Bench HARD \citep{suzgun-etal-2023-challenging}, FrontierMath \citep{frontiermath2024}, or the subject of the current study, GSM8K  \citep{cobbe2021gsm8k} with several descendants, 
including GSM-Symbolic \citep{mirzadeh2025gsm}, GSM-Plus \citep{Li2024gsmplus}, GSM-Identity \citep{negi2026gsmidentity}, and GSM-Ranges \citep{shrestha2025gsmranges}. However, systematic evaluation of LLMs poses a number of challenges, rendering the interpretation and generalisability of the results more nuanced. As demonstrated by \citet{Sclar2023quantifying}, LLMs display high sensitivity to prompt formatting, with surface-level modifications linked to potentially vast changes in performance independently of difficulty of the problem to be solved. The effect seems to be independent of model family or scale \citep{Sclar2023quantifying, alhetelah2026measuring}, with changes as trivial as letter case or separator swap drastically changing models' outputs \cite{errica2025did}. Due to impracticality of manual analysis, benchmarks frequently expect strict formatting of the answers. However, literature suggests that formatting constraints may induce what \citet{cazares2026morecognitiveloadsingleprompt} terms \textit{cognitive load collapse}, leading to not just incorrect answers, but a complete failure to handle the problems \citep{cazares2026morecognitiveloadsingleprompt, mittal2026didforgetiasked}. Furthermore, number processing in LLMs is sensitive to tokenisation mechanisms used \citep{spathis2024first,singh2024tokenization}. Models struggle to generalise their arithmetic capabilities to larger numbers, expressed with more digits \citep{yang2023gpt, zhou2024what} --- an issue resulting in recent argumentation for single-token number embeddings \citep{kreitner2025efficient,zhou2025fone}. Approaches such as Program of Thought \citep{chen2023program} and Program-Aided Language Models \cite{gao2023pal} attempt to circumvent the large-number limitations through instructing models to generate Python code from which the final numerical answer is calculated using an external interpreter. However, as demonstrated by \citeauthor{shrestha2025gsmranges} in their high-number-range adaptation of the GSM8K benchmark, offloading the calculations to an external tool might not be sufficient as higher values frequently deteriorate the underlying reasoning processes of a model.

\section{Methodology}

This section details the experimental setup underlying our re-evaluation. We begin by defining the datasets, metrics, and models used (Sections~\ref{sec:definitions}–-\ref{sec:models}), before describing the prompt formats designed to probe candidate failure modes (Section~\ref{sec:prompt-formats}). We then introduce a large-number effect metric to separate arithmetic difficulty from reasoning failures per se (Section~\ref{sec:number-effect}), and describe the statistical framework — based on GLMMs with per-question random effects — used to assess significance of the observed effects (Section~\ref{sec:glmm}). Finally, Section~\ref{sec:experiments} presents the research questions and experimental design.
The Python code developed for all experiments and evaluations performed in the scope of this project is available at \url{https://github.com/the-mysh/gsm-symbolic-benchmarking}.

\subsection{Definitions}
\label{sec:definitions}
\paragraph{Datasets} Our experiments focus on two datasets from the GSM-Symbolic study \citep{mirzadeh2025gsm}: \textbf{\textit{GSM-Variants}} (labelled as \textit{main} or \textit{Symbolic} in the study), i.e. the 5000 template-generated questions; and \textbf{\textit{GSM-Base}} (\textit{GSM8K 100} in \citeauthor{mirzadeh2025gsm}), i.e. the corresponding 100 originals chosen from \citeauthor{cobbe2021gsm8k}. A question from the \textit{GSM-Base} dataset and its 50 variants from \textit{GSM-Variants} share the same integer
\textit{template ID}. We refer to the phenomenon of decreased accuracy on \textit{GSM-Variants} questions (w.r.t. to performance \textit{GSM-Base} problems) as the \textbf{\textit{variant effect}}.

\paragraph{Metrics}
Across the paper, we express average accuracy as a percentage of correctly answered questions. Accuracy deltas (differences in accuracy between two datasets) are expressed in percentage points (pp).
We use the term \textbf{\textit{variant performance delta}} to describe raw change in accuracy, in percentage points, from \textit{GSM-Base} to \textit{GSM-Variants} datasets: 
$\Delta_{var} = \overline{A}_{GSM-Variants} - \overline{A}_{GSM-Base}$. The accuracies ($\overline{A}$)
in this context are averaged across the entire respective datasets.

\subsection{Models}\label{sec:models}
The GSM-Symbolic study evaluated the performance of 25 LLMs. For economical and practical reasons, we exclude the four proprietary models (all developed by OpenAI) from our analysis. In addition, we omit Phi-3-small-128k-instruct (Microsoft) due to technical difficulties. Otherwise, we follow the original list, presenting 20 open-weight models for our experiments. The models and limitations can be consulted in Appendix~\ref{apx:models}.

\subsection{Prompt Formats and Failure Modes}\label{sec:prompt-formats} 
Throughout the work and in accordance with the GSM-Symbolic study, we use 8-shot prompting with the standard GSM8K shots \citep{eval-harness}, where a shot is a single mathematical problem with a step-by-step solution. First, we reproduce the GSM-Symbolic results using the original prompt format, which we refer to as \textit{GSM prompt}. In later experiments (see Section~\ref{sec:experiments}), we modify the solutions to the eight example problems alongside with the prompt introduction line. The exact templates of all prompts used in our experiments, described below, are included in Appendix~\ref{apx:full-prompts}.

To test whether the models indeed rely on the exact phrasing of the original \textit{GSM prompt}, we propose a simple rewording of the original answers. We remove the explicit instruction to follow a stepwise approach, but include step introduction phrases (\textit{First, calculate...}, \textit{Next, ...}, etc.) where their relevance is justified by the problem structure. We refer to this prompt format as \textbf{\textit{simple natural language (NL) prompt}}.

Solving grade school problems exemplified by GSM8K requires robust variable binding\footnote{\textit{The binding problem} in cognitive science describes the set of mechanisms by which the brain associates specific attributes, such as value or colour, with (abstract) objects maintained in the working memory \cite{roskies1999binding}.} and internal state tracking. To isolate the fragility of these capabilities as the possible factors contributing to the variant effect, we propose a \textbf{\textit{structured NL prompt}}. In addition to the modified phrasing of the \textit{simple NL prompt}, we include explicit sections with headers: \textit{Given}, \textit{To find}, and \textit{Solution}, detailing all the relevant quantities in the first section and re-stating the calculation target in the second. By introducing these more rigid formatting constraints, we aim to provide a supportive scaffolding, partially offloading the cognitive capabilities to the introduced structure. On the other hand, as discussed before, strict formatting requirements could be a factor negatively influencing performance of some models \citep{cazares2026morecognitiveloadsingleprompt, mittal2026didforgetiasked}.

Finally, we adapt the example answers into two types of \textit{code} prompts, where the model is asked to provide a valid Python function accepting no arguments, performing the necessary calculations using variables defined inside the function, and returning the calculated result. With these prompts, we aim to investigate arithmetic processing as the potential source of the variant effect. Analogously to the NL prompts described above, we propose two code prompts: a \textbf{\textit{structured code prompt}} following the same three-part structure as the \textit{structured NL prompt}, and a \textbf{\textit{simple code prompt}}, which maximally limits any additions over the standard short explanations of steps taken to solve the problems.

Answers from experiments using the \textit{GSM}, \textit{simple NL}, and \textit{structured NL} prompts are extracted using standard pattern matching: we first attempt to match the target response line (\textit{The final answer is}) is followed by a number; if this fails, we extract the last number in the response, following GSM-Symbolic. For the code prompts, we attempt to execute the output function via a Python interpreter after safety checks, collecting error type information for non-executing outputs (Appendix~\ref{apx:error-types}).

\subsection{The Effect of Large Numbers}\label{sec:number-effect}
We believe it is imperative to separate purely arithmetic difficulties from limitations of mathematical reasoning understood as a process of constructing logically consistent solution procedures, grounded in context and common sense.
To this end, we evaluate whether larger numbers in problems' texts are a predictor of altered performance.
We construct a \textit{large-number effect metric}, denoted as $\gamma$, reflecting the total digit load of a question. For each question, we extract all numerically expressed integers (i.e. we ignore cases such as 'three', 'twice', 'half', and all numerically expressed fractions). We then compute logarithms of base 10 for each number (negative numbers, if any, are taken at their absolute value; for 0, we take the value of 1) and sum the obtained values for a given question. For a given question,
\begin{equation}
    \gamma = \sum_{n \in \mathcal{N}} log(n)
\end{equation}
where $\mathcal{N}$ is the set of all numerically expressed integers in the question text. This metric captures two factors: firstly, the logarithm assigns unit growth for each added digit, accounting for the number-splitting effect in tokenisation; secondly, larger numbers are expected to appear less in training data and therefore errors in arithmetic operations involving these numbers can be deemed more likely.

We test for this \textbf{\textit{large-number effect}} on combined data from both \textit{GSM-Variants} and \textit{GSM-Base} datasets, separately for each LLM. We assess two aspects: (1) whether the large-number effect is present (i.e. whether larger numbers are associated with decreased performance); (2) whether, controlling for the number magnitude, variant effect remains significant. The details of the test setup are discussed in Section~\ref{sec:glmm}. When fitting the statistical model, we centre $\gamma$ to zero mean; we denote the centred metric as $\gamma_c$.

\subsection{Statistical Significance Testing}\label{sec:glmm}
We model the influence of the different factors on response accuracy employing GLMMs. We use a binomial family with logit link, appropriate for the binary correct/incorrect outcome. 
The choice of the model type (GLMM) allows accounting for the non-independence of repeated observations --- results for the same question template should not be treated as statistically independent.

We define two instances of the model. With \textbf{\textit{GLMM 1}}, we model the binary outcome $Correct$ --- whether a given question was solved correctly by a given LLM --- as a function of question type ($Variant$: 0 for GSM-Base, 1 for GSM-Variants) as a fixed effect, with a random intercept (baseline difficulty) per template ($Id$). The random intercept accounts for the fact that GSM-Base questions and their GSM-Variants share the same underlying template, making their responses non-independent; it allows the baseline difficulty to vary freely across templates rather than assuming all questions are equally hard. The fixed effect of $Variant$ then estimates whether belonging to the variant set is associated with a reliable change in accuracy, over and above this template-level variability.
In Wilkinson notation \cite{wilkinson1973symbolic, Bates2015lme4}, GLMM 1 is specified as follows:
\begin{equation}
    Correct \sim Variant + (1 | Id)
\end{equation}

Next, we extend GLMM 1 by including the centred large-number effect metric $\gamma_c$ (see Section~\ref{sec:number-effect}) as a second fixed effect. This yields \textbf{\textit{GLMM 2}}:
\begin{equation}
    Correct \sim Variant + \gamma_c + (1 | Id)
\end{equation}
where the estimate for $\gamma_c$ tests whether numeric load is independently associated with accuracy, and the estimate for $Variant$ --- whether the variant effect persists after controlling for it, since questions from GSM-Variants may differ from their GSM-Base counterparts not only in surface form but in the numerical values they employ. We fit both models separately for each LLM (see Section~\ref{sec:models}) and prompt format (see Section~\ref{sec:prompt-formats}).
Initially, we performed a single fit for each case (Wald approach). However, facing several non-convergent fits and prompted by our reviewers to run additional checks, we adopted bootstrapping over the previously defined GLMMs as our primary inferential method throughout.\footnote{We found that Wald- and bootstrap-derived significance calls disagreed in several cases, not only for non-convergent fits but also for some fits that converged without warning. Detailed analysis available in Appendix \ref{apx:degenerate-fit}.} This methodological switch both circumvents the non-convergence issues (a vast majority of resampled fits converges cleanly) and avoids the normality assumptions inherent to Wald statistics. We use 2000 resamples per model-prompt format combination, resampling question templates (not individual observations) with replacement to preserve the hierarchical structure of the original model. We perform the fitting using a Python library \texttt{pymer4} \cite{jolly2018pymer4}, wrapping over R's LME4 \cite{Bates2015lme4}. 

Processing the bootstrap results, we discard any individual estimates raising convergence warnings; we report the bootstrap success rates in supplementary results tables throughout Appendix \ref{apx:full-results}.
We present the obtained fixed-effect estimates as odds ratios (OR) derived as the median (computed on log-OR scale and subsequently converted to linear OR) of clean bootstrap estimates; we choose the median over mean for better robustness to non-normality. OR > 1 indicates increased odds of a correct response relative to baseline, OR < 1 --- decreased odds. In figures, we report log ORs for visual symmetry around zero. The P values corresponding to the estimates serve as an indication of statistical significance of each of the effects. The P values are derived as the twice the smaller of the two bootstrap tail probabilities, i.e. the proportion of clean bootstrap estimates falling on the opposite side of 0 than the majority of the mass:
\begin{equation}
    p=2\cdot min(\frac{1}{B}\sum_{b=1}^B\mathbbm{1}_{\theta_b\leq0}, \frac{1}{B}\sum_{b=1}^B\mathbbm{1}_{\theta_b\geq0})
\end{equation}
where $B$: number of cleanly converged bootstrap estimates; $\theta_b$: b-th bootstrap estimate on the log-odds scale. Consistently with the above, we derive the 95\% confidence intervals bounds on the estimates as the 2.5-th and 97.5-th percentile of clean estimates.

Note: to evaluate the global significance of the variant and large-number effects across all models, we apply the Holm-Bonferroni correction across the evaluated LLMs, separately per effect. We select this over the standard Bonferroni procedure as it offers superior statistical power while maintaining strict control over the family-wise error rate, ensuring that our critique does not unfairly penalize marginal but real effects. We nonetheless report and discuss uncorrected P values alongside corrected ones, as our primary interest lies in model-specific failure profiles rather than family-wise conclusions.

\subsection{Research Questions and Experiments}\label{sec:experiments}
We consider three principal research questions.

\textbf{RQ1.} Does the variant effect reported by the GSM-Symbolic study exhibit statistical significance under GLMM?

\textbf{RQ2.} What are the possible failure modes responsible for the observed performance decline?

\textbf{RQ3.} Are models sensitive to larger numbers in the range used in the \textit{GSM-Base} and \textit{GSM-Variants} datasets of GSM-Symbolic?

To answer RQ1, we reproduce the GSM-Symbolic experiments (Ex1; model limitations in Section~\ref{sec:models}), fitting GLMM 1 (Section~\ref{sec:glmm}) per LLM and comparing P values against $\alpha = 0.05$ to evaluate statistical significance of the variant effect.

For RQ2, models significant under Ex1 are evaluated on the four alternative prompt formats (Ex2.A–D: \textit{simple NL}, \textit{structured NL}, \textit{simple code}, and \textit{structured code} prompts --- see Section~\ref{sec:prompt-formats}), using GLMM 1 per LLM and prompt format to assess whether the variant effect vanishes.

For RQ3, we reuse raw model evaluation data from Ex2.A to Ex2.D in corresponding experiments Ex3.A to Ex3.D. We fit GLMM 2 (see Section~\ref{sec:glmm}) to combined \textit{GSM-Variants} and \textit{GSM-Base} results (per LLM and prompt format) to evaluate the statistical significance of the large-number effect, as well as the residual significance of large-number-effect-corrected variant effect.

\section{Results}
In this section, we present the results of our experiments: evaluating statistical significance of results presented in \citet{mirzadeh2025gsm} and investigating possible underlying causes of the reported performance decline.

\subsection{Reproduction of GSM-Symbolic}\label{sec:results-gsm}

\begin{figure*}[!ht]
  \includegraphics[width=\linewidth]{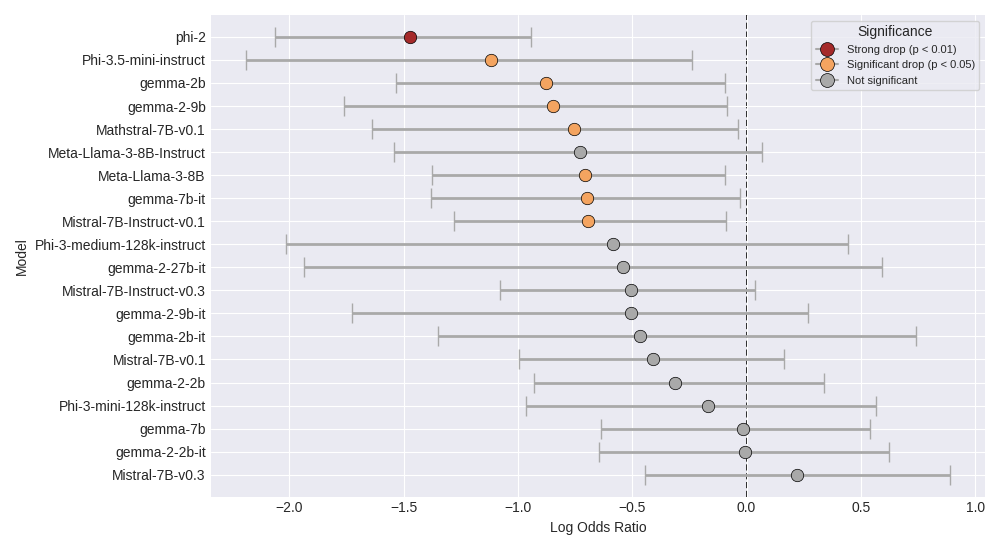}
  \includegraphics[width=\linewidth]{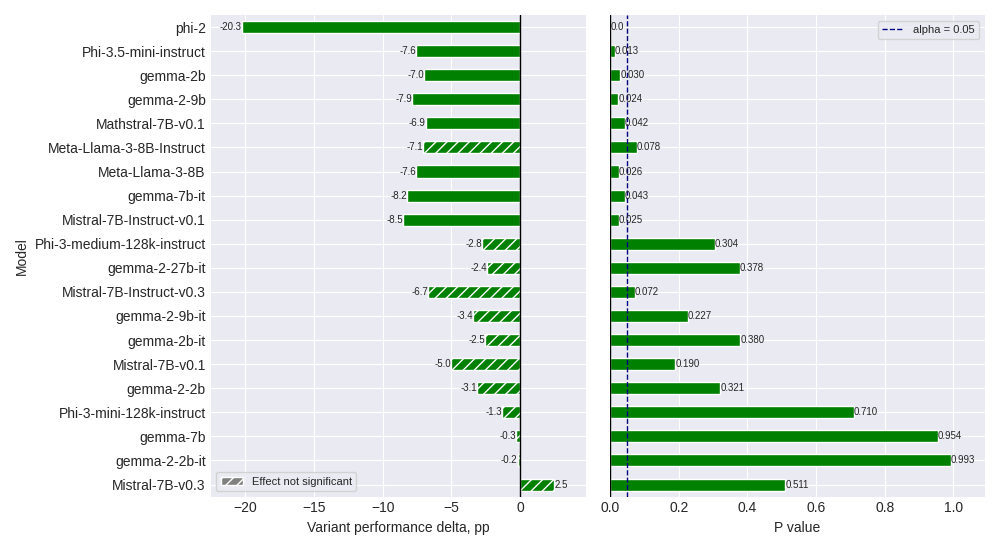}
  \caption{Reproduction of the GSM-Symbolic benchmark. \plotdesc}
  \label{fig:standard_full}
\end{figure*}

Results of Ex1 are illustrated in Figure~\ref{fig:standard_full}. Full details are included in Tables~\ref{tab:GSM-results} and \ref{tab:GSM-stats} in Appendix~\ref{apx:full-results}. Out of 20 models subjected to GLMM evaluation, 8 show statistically significant variant effect when considered individually.\footnote{Our initial Wald approach yielded statistically significant results for two additional models; for one of these, the Wald GLMM fit was degenerate. For more details, please refer to Appendix~\ref{apx:degenerate-fit}.}

\subsection{Alternative Prompt Formats}\label{sec:results-prompts}
\begin{figure*}[!ht]
  \includegraphics[width=\linewidth]{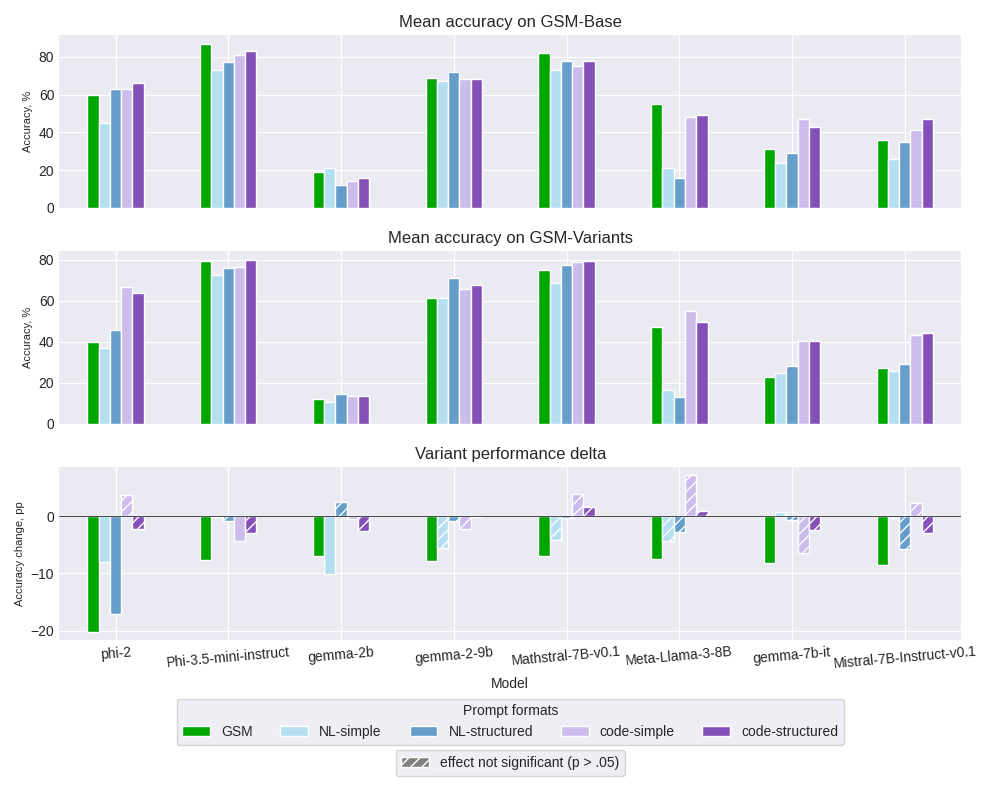}
  \caption{Accuracies and and accuracy changes (variant performance delta $\Delta_{var}$, see Section~\ref{sec:definitions}) across the five prompt formats (see Section~\ref{sec:prompt-formats}), for eight example LLMs (see Section~\ref{sec:results-prompts}). Hatching in the bottom plot indicates the given variant effect is not significant (p~>~0.05, prior to the Holm-Bonferroni correction).}
  \label{fig:prompts-variant-effect}
\end{figure*}

Detailed results for the four alternative prompt formats (Ex2.A--Ex2.D) are shown in Appendix~\ref{apx:prompt-formats-results}. For five of the eight LLMs, the variant effect is rendered statistically insignificant on all four alternative prompt formats. These models are: gemma-2-9b, Mathstral-7B-v0.1, Meta-Llama-3-8B, gemma-7b-it, and Mistral-7B-Instruct-v0.1. Significance is still observed for phi-2 (with both NL prompts), gemma-2b (with \textit{simple NL prompt}, and Phi-3.5-mini-instruct (with \textit{simple code prompt}). In the latter case, the P value of 0.049526 barely crosses the significance threshold. The dynamics of the variant effect for the eight considered models are shown in a Figure~\ref{fig:prompts-variant-effect}.

\subsection{The Large-Number Effect}\label{sec:number-effect-results}
\begin{figure*}[!ht]
  \centering
  \includegraphics[width=\linewidth]{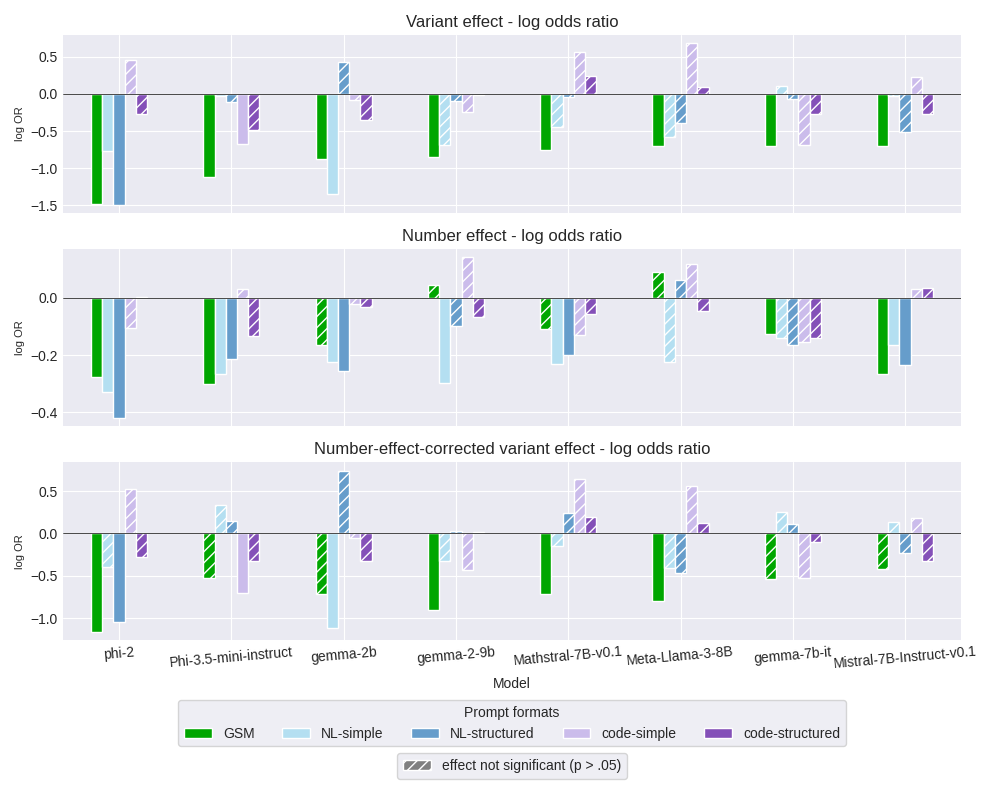}
  \caption{Variant effect and large-number effect --- a combined case study of eight LLMs. Top: log-odds ratio for raw variant effect (performance change on \textit{GSM-Variants} vs \textit{GSM-Symbolic}; $Variant$ variable in GLMM 1). Middle: log-odds ratio for large-number effect ($\gamma_c$ in GLMM 2). Bottom: log-odds ratio for variant effect corrected for the large-number effect ($Variant$ in GLMM 2). Hatching indicates the respective effect is not significant (p~>~0.05, uncorrected).}
  \label{fig:prompts-number-effect}
\end{figure*}

Figure~\ref{fig:numbers} in Appendix~\ref{apx:number-effect} presents the distributions of numerically expressed integers extracted from questions of the \textit{GSM-Base} and \textit{GSM-Variants} dataset of the GSM-Symbolic benchmark. To formally confirm the visible systematic shift towards larger numbers in \textit{GSM-Variants}, we conduct a two-sample Kolmogorov-Smirnov test on the unrolled occurrences in both datasets. The test confirms a highly statistically significant difference between the underlying distributions ($D = 0.1201$, $p < 0.001$). This result contradicts the cursory assessment made in one of the ablation studies of the GSM-Symbolic paper (\textit{Ablation: The Impact of Arithmetic Accuracy}, Appendix A.6), where the authors stated: ``\textit{overall, the range of numbers does not increase significantly}'' \citep{mirzadeh2025gsm}.

The results of experiments E3.A to Ex3.D, i.e. of our evaluation of the large-number effect on model performance, are shown in Figure~\ref{fig:prompts-number-effect} as well as Table~\ref{tab:number_effect_odds} in Appendix~\ref{apx:number-effect}. As demonstrated by these results, large numbers are a predictor of decreased performance on pooled \textit{GSM-Base} and \textit{GSM-Variants} questions for four of the eight models with the canonical \textit{GSM prompt}; for 3 of these (and for one additional model where the number effect falls short of significance), correcting for the large-number effect (GLMM~2) renders the variant effect not statistically significant anymore. Number effect is further significant for 6 models with the \textit{simple NL prompt} and 5 models with the \textit{structured NL prompt}.
Either of the code prompts removes the significance of the large-number effect. The large-number-corrected variant effect remains significant only for three cases across the eight models and the four alternative prompts: for gemma-2b with the \textit{simple NL prompt}, phi-2 with the \textit{structured NL prompt}, and Phi-3.5-mini-instruct for the \textit{simple code prompt}.

\section{Discussion}
We discuss the implications of our findings concerning the statistical standards of LLM evaluation and model-specific failure profiles.

\subsection{The Importance of Statistics}
\citeauthor{mirzadeh2025gsm} argue that the observed drop in performance on template versions GSM8K problems indicates lack of genuine reasoning capabilities in the affected models and by extension, in LLMs of the time. Our results suggest that this conclusion might be premature. The variant effect is strikingly less prevalent a phenomenon than implied in the GSM-Symbolic study. On the contrary, we demonstrate that only 8 of the 20 models we re-evaluate display statistically significant variant effect. 

Furthermore, our analysis based on the models which individually (pre-correction) exhibit significant variant effect suggests that the influence of number distribution shift should not be ignored when interpreting results of GSM-Symbolic. While sensitivity to large numbers is indubitably an issue to be considered, failures due to arithmetic capabilities should not, in our opinion, be classified as reasoning errors.

\subsection{Case Study of Failure Modes}
While the variant effect may be indicative of data contamination in the affected models, our results suggest that the underlying failure mechanisms vary. Our analysis reveals several likely phenomena --- behaviourally-motivated hypotheses subject to further verification through mechanistic interpretability techniques --- described below.

\paragraph{Fragility of variable binding}
For gemma-2b, the variant effect remains significant under simple rewording of the example solutions (\textit{simple NL prompt}), but vanishes when the \textit{structured NL prompt} is used. This behaviour fits our earlier discussed assumption that the supportive scaffolding introduced by this prompt offloads the difficult (for the model) tasks of associating values with objects and tracking the resulting internal state to the external scratch pad of the generated response.

\paragraph{Limited arithmetic capabilities}
For phi-2, variant effect remains significant across the NL prompt adaptations, but disappears entirely with the code prompts, where the otherwise significant large-number effect also vanishes. The behaviour of this model suggests limitations in performing arithmetic operations with larger numbers; its reasoning capabilities (for the purpose of the current work understood as the ability to solve grade school mathematical problems) seem to remain largely intact through meaning-preserving prompt modifications.

\paragraph{Reliance on learnt patterns}
Models: gemma-2-9b and Mathstral-7B-v0.1 exhibit the variant effect only with the original \textit{GSM prompt}, but unlike other studied models, are fully or partly immune to the number effect --- possibly owing to the larger capacity facilitated by the higher parameter count (Gemma with 9B parameters) or training specific for mathematical reasoning (Mathstral). For these two models, we do not find an explanation for the original variant effect other than replication of patterns learnt from training data.

\paragraph{Dual-task interference}
The indifference to large numbers is also observed for Meta-Llama-3-8B. However, here we notice additional patterns: NL perturbations (\textit{simple NL prompt}, \textit{structured NL prompt}) are associated with a strong decrease in accuracy and, as shown in Fig.~\ref{fig:nl-errors}, a large number of empty answers. This behaviour can be interpreted in terms of \textit{dual-task inference}, mirroring a phenomenon known in psychology \cite{pashler1994dual}. Attempting to follow out-of-distribution formatting requirements while simultaneously solving a mathematical problem, the model fails at both tasks. 

\paragraph{}Overall, our results demonstrate the diversity of possible root causes for failures on reasoning benchmarks and highlight the inappropriateness of blanket statements when analysing the outcomes.

\section{Conclusions}
This work presents a re-evaluation of the GSM-Symbolic benchmark, and its primary message is methodological: claims about the reasoning capabilities --- or incapabilities --- of LLMs demand statistical rigour. The widespread absence of significance testing in LLM evaluation is not merely an academic concern; it shapes research agendas, informs policy, and influences public understanding of AI systems.

Our results challenge the central claim of \citet{mirzadeh2025gsm}. Applying bootstrapped GLMM with appropriate random effects for item-level variability, we find that the variant effect is statistically significant for only 8 of 20 re-evaluated models. For the models showing significance pre-correction, we demonstrate that a systematic shift towards larger integers in the main GSM-Symbolic dataset constitutes a previously overlooked confound, partially or fully explaining the decline in several cases.

Crucially, for models whose variant effect proves robust to prompt perturbations, no single failure mode prevails. Instead, we observe a diverse behavioural landscape: limitations in arithmetic capabilities, fragile variable binding, and dual-task interference --- the compounding of out-of-distribution entity processing with rigid structural constraints. Only two of the LLMs seem to purely rely on patterns learnt from their training data. The heterogeneity of the results underscores the importance of careful, case-based investigation, possibly supported by mechanistic understanding, in evaluating or speculating about cognitive capabilities of LLMs.
We hope this work contributes to a more epistemically careful discourse around LLM evaluation, one that treats statistical validation not as an optional refinement, but as a prerequisite for credible claims.


\section*{Limitations}

While our re-evaluation provides a statistically rigorous perspective on the GSM-Symbolic benchmark, our methodology has several constraints that invite further investigation.

\paragraph{Simplification of the large-number effect metric} The metric we introduce to quantify arithmetic difficulty ($\gamma$, see Section~\ref{sec:number-effect}) acts as a proxy for total number load. However, it only captures the magnitude of integers present in questions. It does not account for the presence of fractions, decimals, the complexity of specific arithmetic operations (e.g., carrying over in addition, division by primes), or the nuances of model-specific tokenisation boundaries, all of which are known to impact arithmetic performance. Our chosen heuristics offers a balance between capturing two major expected confounds and the overall clarity of presentation (i.e. a single, expressive metric for the discussed class of numerical issues). Further work could devise individual metrics isolating the possible contributing effects. 

\paragraph{Granularity of prompt perturbations} For five models, the variant effect vanishes under all tested prompt variations. While we interpret this as a partial reliance on superficial pattern matching, we cannot rule out extreme sensitivity to formatting. Further studies utilizing small-scale perturbations are required to isolate pattern matching from general format fragility.

\paragraph{Behavioural abstraction of cognitive mechanisms} Our hypotheses regarding dual-task interference and the fragility of variable binding are inferred from black-box behavioural evaluations. While these cognitive aspects align with the observed performance deltas across prompt formats, validating the exact root causes of these failures would require white-box interpretability techniques, such as activation patching or circuit tracing. We also acknowledge that the observed behavioural patterns might reflect the representation of the particular prompt formats in pretraining data rather than genuine competences of the models. A separate question, which we do not attempt to tackle in this work, is to what extent the ability to solve mathematical grade school problems can be treated as a proxy for reasoning capabilities.

\paragraph{Scope of evaluated models} For practical and economic reasons, our evaluation is restricted to open-weight models, all under 30 billion parameters. We excluded frontier proprietary models. Moreover, with reproduction of the GSM-Symbolic study as our primary motivation, we do not include more recent LLMs. Consequently, our findings regarding format sensitivity and the statistical insignificance of the variant effect may not fully generalise to the state-of-the-art models that typically define upper-bound reasoning capabilities. However, the methodological critique, emphasising the necessity of statistical significance testing with appropriate tools and of controlling for dataset distribution shifts, applies universally to all models.

\section*{Ethical Considerations}

\paragraph{Epistemological responsibility and AI discourse}
This work challenges definitive claims regarding the presence or absence of ``genuine reasoning'' capabilities in Large Language Models. We emphasise the ethical necessity of rigorous statistical validation before making broad assertions about AI capabilities. Prematurely concluding that models lack reasoning abilities  --- or conversely, overestimating their abilities based on contaminated benchmarks --- can misinform public understanding, guide flawed policy decisions, and skew future safety research. Our findings advocate for a more nuanced, statistically grounded discourse regarding model evaluation.

\paragraph{Use of cognitive metaphors}
Throughout this paper, we employ terminology derived from cognitive science and psychology, such as ``cognitive load'' and ``dual-task interference''. We explicitly state that these terms are used as functional metaphors to describe mechanistic and architectural phenomena within LLMs, such as variable binding and internal state tracking. Their use is intended to provide an intuitive framework for understanding failure modes, and should not be interpreted as an assertion that these models possess human-like sentience, biological cognition, or consciousness.

\paragraph{Computational and environmental cost}
We acknowledge the environmental impact associated with our experimental methodology. Re-evaluating the GSM-Symbolic benchmark required running inference across 20 distinct LLMs on thousands of prompt variations. While this incurs a notable compute and carbon footprint, we believe this cost is justified by the critical need to validate high-impact benchmark results and prevent a potential replication crisis within the Natural Language Processing community.

\paragraph{Use of AI assistance}
In this work, we used the assistance of AI chat bots (Claude, Gemini) to identify the relevant sources cited across the article and in particular those mentioned in the Related Work section, as well as to help with concise phrasing of the text. In addition, coding agents (GitHub Copilot) were employed to help construct model evaluation, data analysis, and visualisation scripts.

\section*{Acknowledgements}
Should not be included in the review version

The authors would like to express their appreciation towards the anonymous reviewers of the May 2026 ARR cycle, whose thoughtful suggestions meaningfully helped strengthen the paper.

Díaz acknowledges TSI-100927-2023-1 Project (Transformation and Resilience Plan from the EU NextGen through the Ministry for Digital Transformation and the Civil Service), Google Research Scholar and Grants PID2023-149128NB-I00 and PID2023-150070NB-I00 funded by MICIU/AEI /10.13039/501100011033 and by ERDF, EU.

This work was supported by the Center for Responsible AI project with reference C628696807-00454142, financed by the Recovery and Resilience Plan, and by national funds through Fundação para a Ciência e a Tecnologia, I.P. (FCT) under projects UID/50021/2025 and UID/PRR/50021/2025.

\bibliography{custom}


\section*{Appendix}
\appendix


\section{Artifacts}
\subsection{Software and Infrastructure}
Model inference and evaluation pipelines were implemented in Python utilizing the PyTorch framework \citep{paszke2019pytorch} and the Hugging Face \texttt{transformers} library \citep{wolf2020huggingfacestransformersstateoftheartnatural}, alongside \texttt{datasets} \citep{lhoest-etal-2021-datasets} for benchmark data management. Data manipulation and preprocessing were conducted using \texttt{pandas} \cite{mckinney-proc-scipy-2010} and \texttt{numpy} \cite{harris2020array}.

All statistical hypothesis testing, including the formulation of GLMMs, was executed using \texttt{pymer4} \cite{jolly2018pymer4} --- which interfaces with the R \texttt{lme4} package \cite{Bates2015lme4} --- as well as \texttt{scipy} \cite{Virtanen_2020} and \texttt{statsmodels} \cite{seabold2010statsmodels}. Visualizations were generated relying on \texttt{matplotlib} \cite{Hunter:2007} and \texttt{seaborn} \cite{Waskom2021}.

\subsection{Computational Experiments}
The evaluations were performed on NVIDIA A40 and NVIDIA V100 GPUs. The total computational budget required to run the full evaluation pipeline across all 20 models and prompt variations is estimated at approximately 1,000 GPU hours. The evaluated open-weight models range in size from 2 billion to 27 billion parameters, with exact model versions and sizes specified in Appendix~\ref{apx:models}.

\subsection{Artifact Licences and Terms of Use}
The artifacts utilized in this study are publicly available for research purposes. The GSM8K dataset \citep{cobbe2021gsm8k} is distributed under the MIT licence and the GSM-Symbolic benchmark \citep{mirzadeh2025gsm} --- under CC-BY-NC-ND 4.0. The evaluated Large Language Models are subject to their respective open-weight licenses: models from the Mistral and Mathstral families are released under the Apache 2.0 License; the Microsoft Phi models are released under the MIT License; the Google Gemma models are governed by the Gemma Terms of Use; and the Meta-Llama-3 models are governed by the custom Meta Llama 3 Community License. All third-party software packages and libraries used for inference and statistical analysis (e.g., PyTorch, Hugging Face transformers, pymer4) are open-source and utilized in accordance with their respective permissive licenses (primarily MIT, BSD, and Apache 2.0). All models, datasets, and software were used strictly for research and evaluation purposes. The custom evaluation code and data analysis scripts developed for this study will be released under the MIT License upon publication to facilitate reproducibility.

\section{Model Specifications}\label{apx:models}
In our experiments, we use models downloaded from the Hugging Face repository \citep{wolf2020huggingfacestransformersstateoftheartnatural}.
Table \ref{tab:models} lists the models we evaluate together with the corresponding checkpoints (commit ids) for reproducibility.  When reporting the results throughout the paper, we omit model family names for brevity.

\begin{table*}[t]
\small
\centering
\begin{tabular}{lcc}
\toprule
Model & Commit ID & Parameters \\
\midrule
google/gemma-2b & 9cf48e52b224239de00d483ec8eb84fb8d0f3a3a & 2B \\
google/gemma-2b-it & 96988410cbdaeb8d5093d1ebdc5a8fb563e02bad & 2B \\
google/gemma-7b & ff6768d9368919a1f025a54f9f5aa0ee591730bb & 7B \\
google/gemma-7b-it & 9c5798d27f588501ce1e108079d2a19e4c3a2353 & 7B \\
google/gemma-2-2b & c5ebcd40d208330abc697524c919956e692655cf & 2B \\
google/gemma-2-2b-it & 299a8560bedf22ed1c72a8a11e7dce4a7f9f51f8 & 2B \\
google/gemma-2-9b & 33c193028431c2fde6c6e51f29e6f17b60cbfac6 & 9B \\
google/gemma-2-9b-it & 11c9b309abf73637e4b6f9a3fa1e92e615547819 & 9B \\
google/gemma-2-27b-it & aaf20e6b9f4c0fcf043f6fb2a2068419086d77b0 & 27B \\
microsoft/phi-2 & 810d367871c1d460086d9f82db8696f2e0a0fcd0 & 2.7B \\
microsoft/Phi-3-mini-128k-instruct & f3c06aed622e14ca0abf5115094e4fc9a9948f36 & 3.8B \\
microsoft/Phi-3-medium-128k-instruct & a088b37c71d441ab6d862bb3fcfe6165b3014702 & 14B \\
microsoft/Phi-3.5-mini-instruct & 2fe192450127e6a83f7441aef6e3ca586c338b77 & 3.8B \\
mistralai/Mistral-7B-v0.1 & 27d67f1b5f57dc0953326b2601d68371d40ea8da & 7B \\
mistralai/Mistral-7B-Instruct-v0.1 & ec5deb64f2c6e6fa90c1abf74a91d5c93a9669ca & 7B \\
mistralai/Mistral-7B-v0.3 & caa1feb0e54d415e2df31207e5f4e273e33509b1 & 7B \\
mistralai/Mistral-7B-Instruct-v0.3 & c170c708c41dac9275d15a8fff4eca08d52bab71 & 7B \\
mistralai/Mathstral-7B-v0.1 & ec3a48484ef241dfe03282edcb0f25e564923823 & 7B \\
meta-llama/Meta-Llama-3-8B & 8cde5ca8380496c9a6cc7ef3a8b46a0372a1d920 & 8B \\
meta-llama/Meta-Llama-3-8B-Instruct & 8afb486c1db24fe5011ec46dfbe5b5dccdb575c2 & 8B \\
\bottomrule
\end{tabular}
\caption{Models used in the experiments with Hugging Face checkpoints (commit ids) and parameter counts.}
\label{tab:models}
\end{table*}

Models from the GSM-Symbolic paper which were omitted in the current study:
\begin{itemize}
    \setlength{\itemsep}{-0.5em}
    \item microsoft/Phi-3-small-128k-instruct
    \item gpt-4o-mini
    \item gpt-4o
    \item o1-mini
    \item o1-preview
\end{itemize}

We did not evaluate closed models (GPT-4o, o1) because of technical and economical constraints. For microsoft/Phi-3-small-128k-instruct, configuring Flash Attention on our hardware proved impassable despite a considerable time investment.

\onecolumn

\newpage
\section{Prompt Formats}
In this section, we present the full prompt formats used in our experiments. See Section~\ref{sec:prompt-formats} for more details.

\label{apx:full-prompts}

\subsection*{GSM Prompt}
To facilitate comparison with the original prompt format from GSM8K and GSM-Symbolic studies, we reproduce here the full 8-shot \textit{GSM prompt}, as specified in EleutherAI LM-Harness framework \citep{eval-harness}. 
\begin{lstlisting}[caption={\textit{GSM prompt}},label={lst:gsm}]
As an expert problem solver, solve step by step the following mathematical questions.

Q: There are 15 trees in the grove. Grove workers will plant trees in the grove today. After they are done, there will be 21 trees. How many trees did the grove workers plant today?
A: Let's think step by step. There are 15 trees originally. Then there were 21 trees after some more were planted. So there must have been 21 - 15 = 6. The final answer is 6.

Q: If there are 3 cars in the parking lot and 2 more cars arrive, how many cars are in the parking lot?
A: Let's think step by step. There are originally 3 cars. 2 more cars arrive. 3 + 2 = 5. The final answer is 5.

Q: Leah had 32 chocolates and her sister had 42. If they ate 35, how many pieces do they have left in total?
A: Let's think step by step. Originally, Leah had 32 chocolates. Her sister had 42. So in total they had 32 + 42 = 74. After eating 35, they had 74 - 35 = 39. The final answer is 39.

Q: Jason had 20 lollipops. He gave Denny some lollipops. Now Jason has 12 lollipops. How many lollipops did Jason give to Denny?
A: Let's think step by step. Jason started with 20 lollipops. Then he had 12 after giving some to Denny. So he gave Denny 20 - 12 = 8. The final answer is 8.

Q: Shawn has five toys. For Christmas, he got two toys each from his mom and dad. How many toys does he have now?
A: Let's think step by step. Shawn started with 5 toys. If he got 2 toys each from his mom and dad, then that is 4 more toys. 5 + 4 = 9. The final answer is 9.

Q: There were nine computers in the server room. Five more computers were installed each day, from monday to thursday. How many computers are now in the server room?
A: Let's think step by step. There were originally 9 computers. For each of 4 days, 5 more computers were added. So 5 * 4 = 20 computers were added. 9 + 20 is 29. The final answer is 29.

Q: Michael had 58 golf balls. On tuesday, he lost 23 golf balls. On wednesday, he lost 2 more. How many golf balls did he have at the end of wednesday?
A: Let's think step by step. Michael started with 58 golf balls. After losing 23 on tuesday, he had 58 - 23 = 35. After losing 2 more, he had 35 - 2 = 33 golf balls. The final answer is 33.

Q: Olivia has $23. She bought five bagels for $3 each. How much money does she have left?
A: Let's think step by step. Olivia had 23 dollars. 5 bagels for 3 dollars each will be 5 x 3 = 15 dollars. So she has 23 - 15 dollars left. 23 - 15 is 8. The final answer is 8.

Q: <Question here>
A: Let's think step by step. 
\end{lstlisting}

\subsection*{Simple NL Prompt}
\begin{lstlisting}[caption={\textit{Simple NL prompt format.}},label={lst:nonformal}]
As an expert problem solver, solve the following mathematical questions.


Q: There are 15 trees in the grove. Grove workers will plant trees in the grove today. After they are done, there will be 21 trees. How many trees did the grove workers plant today?
A: 
Calculate the difference between trees after planting and trees currently in the grove:
21 - 15 = 6

The final answer is 6.


Q: If there are 3 cars in the parking lot and 2 more cars arrive, how many cars are in the parking lot?
A: 
Calculate the sum of cars originally in the parking lot and the arriving cars:
3 + 2 = 5

The final answer is 5.


Q: Leah had 32 chocolates and her sister had 42. If they ate 35, how many pieces do they have left in total?
A: 
First, calculate how many chocolates the sisters originally had in total:
32 + 42 = 74
Next, subtract the total number of chocolates they ate:
74 - 35 = 39

The final answer is 39.


Q: Jason had 20 lollipops. He gave Denny some lollipops. Now Jason has 12 lollipops. How many lollipops did Jason give to Denny?
A: 
Calculate the difference between the lollipops Jason originally had and the lollipops he has now:
20 - 12 = 8

The final answer is 8.


Q: Shawn has five toys. For Christmas, he got two toys each from his mom and dad. How many toys does he have now?
A: 
First, calculate how many toys in total Shawn got for Christmas:
2 * 2 = 4
Next, calculate the sum of the number of toys Shawn originally had and the number of new toys he got:
5 + 4 = 9

The final answer is 9.


Q: There were nine computers in the server room. Five more computers were installed each day, from monday to thursday. How many computers are now in the server room?
A: 
First, calculate how many new computers were installed in total:
4 * 5 = 20
Next, calculate the sum of the number of computers originally in the room and the total number of new computers installed:
9 + 20 = 29

The final answer is 29.


Q: Michael had 58 golf balls. On tuesday, he lost 23 golf balls. On wednesday, he lost 2 more. How many golf balls did he have at the end of wednesday?
A: 
First, calculate the total number of golf balls Michael lost:
23 + 2 = 25
Next, calculate the difference between the original number of golf balls and the number of golf balls lost:
58 - 25 = 33

The final answer is 33.


Q: Olivia has $23. She bought five bagels for $3 each. How much money does she have left?
A: 
First, calculate the total cost of the bagels:
5 * 3 = 15
Next, calculate the difference between Olivia's original money and what she paid for the bagels:
23 - 15 = 8

The final answer is 8.


Q: <Question here>
A: 
\end{lstlisting}

\subsection*{Structured NL Prompt}
\begin{lstlisting}[caption={\textit{Structured NL prompt format.}},label={lst:formal}]
As an expert problem solver, solve the following mathematical questions, detailing the given data, quantity to find, and necessary calculations.


Q: There are 15 trees in the grove. Grove workers will plant trees in the grove today. After they are done, there will be 21 trees. How many trees did the grove workers plant today?
A: 
Given:
Number of trees currently in the grove: 15
Number of trees in the grove after the workers are done planting today: 21

To find: number of trees the grove workers will plant today.

Solution:
Calculate the difference between trees after planting and trees currently in the grove:
21 - 15 = 6

The final answer is 6.


Q: If there are 3 cars in the parking lot and 2 more cars arrive, how many cars are in the parking lot?
A: 
Given:
Number of cars originally in the parking lot: 3
Number of cars arriving in the parking lot: 2
    
To find: total number of cars in the parking lot.

Solution:
Calculate the sum of cars originally in the parking lot and the arriving cars:
3 + 2 = 5

The final answer is 5.


Q: Leah had 32 chocolates and her sister had 42. If they ate 35, how many pieces do they have left in total?
A: 
Given:
Number of chocolates Leah originally had: 32
Number of chocolates Leah's sister originally had: 42
Number of chocolates Leah and her sister ate: 35

To find: total number of chocolates left.

Solution:
First, calculate how many chocolates the sisters originally had in total:
32 + 42 = 74
Next, subtract the total number of chocolates they ate:
74 - 35 = 39

The final answer is 39.


Q: Jason had 20 lollipops. He gave Denny some lollipops. Now Jason has 12 lollipops. How many lollipops did Jason give to Denny?
A: 
Given:
Number of lollipops Jason had originally: 20
Number of lollipops Jason has left: 12

To find: number of lollipops Jason gave to Denny.

Solution:
Calculate the difference between the lollipops Jason originally had and the lollipops he has now:
20 - 12 = 8

The final answer is 8.


Q: Shawn has five toys. For Christmas, he got two toys each from his mom and dad. How many toys does he have now?
A: 
Given:
Number of toys Shawn originally has: 5
Number of toys his mom and his dad each gave him for Christmas: 2

To find: total number of toys Shawn has now.

Solution:
First, calculate how many toys in total Shawn got for Christmas:
2 * 2 = 4
Next, calculate the sum of the number of toys Shawn originally had and the number of new toys he got:
5 + 4 = 9

The final answer is 9.


Q: There were nine computers in the server room. Five more computers were installed each day, from monday to thursday. How many computers are now in the server room?
A: 
Given:
Number of computers originally in the server room: 9
Number of new computers installed each day: 5
Number of days from monday to thursday: 4

To find: total number of computers in the server room now.

Solution:
First, calculate how many new computers were installed in total:
4 * 5 = 20
Next, calculate the sum of the number of computers originally in the room and the total number of new computers installed:
9 + 20 = 29

The final answer is 29.


Q: Michael had 58 golf balls. On tuesday, he lost 23 golf balls. On wednesday, he lost 2 more. How many golf balls did he have at the end of wednesday?
A: 
Given:
Number of golf balls Michael originally had: 58
Number of golf balls lost on tuesday: 23
Number of golf balls lost on wednesday: 2

To find: number of golf balls left at the end of wednesday.

Solution:
First, calculate the total number of golf balls Michael lost:
23 + 2 = 25
Next, calculate the difference between the original number of golf balls and the number of golf balls lost:
58 - 25 = 33

The final answer is 33.


Q: Olivia has $23. She bought five bagels for $3 each. How much money does she have left?
A: 
Given:
Number of dollars Olivia had originally: 23
Number of bagels Olivia bought: 5
Number of dollars each bagel cost: 3

To find: how many dollars Olivia has left.

Solution:
First, calculate the total cost of the bagels:
5 * 3 = 15
Next, calculate the difference between Olivia's original money and what she paid for the bagels:
23 - 15 = 8

The final answer is 8.


Q: <Question here>
A: 
\end{lstlisting}

\subsection*{Simple Code Prompt}
\begin{lstlisting}[caption={\textit{Simple code prompt format.}}, label={lst:code-short}]
As an expert problem solver and Python developer, write Python functions returning numerical solutions to the following mathematical questions.

Q: There are 15 trees in the grove. Grove workers will plant trees in the grove today. After they are done, there will be 21 trees. How many trees did the grove workers plant today?

A:
def solution():
    trees_now = 15
    trees_after_planting = 21

    # calculate the difference between trees after planting and trees originally
    trees_to_plant_today = trees_after_planting - trees_now
    return trees_to_plant_today


Q: If there are 3 cars in the parking lot and 2 more cars arrive, how many cars are in the parking lot?

A:
def solution():
    cars_before = 3
    cars_arriving = 2

    # calculate the sum of cars originally in the parking lot and the arriving cars
    cars_after = cars_before + cars_arriving
    return cars_after


Q: Leah had 32 chocolates and her sister had 42. If they ate 35, how many pieces do they have left in total?

A:
def solution():
    chocolates_leah_before = 32
    chocolates_sister_before = 42
    chocolates_eaten = 35

    # first, calculate how many chocolates the sisters originally had in total
    chocolates_total_before = chocolates_leah_before + chocolates_sister_before
    # next, subtract the total number of chocolates they ate
    chocolates_left = chocolates_total_before - chocolates_eaten
    return chocolates_left


Q: Jason had 20 lollipops. He gave Denny some lollipops. Now Jason has 12 lollipops. How many lollipops did Jason give to Denny?

A:
def solution():
    lollipops_jason_before = 20
    lollipops_jason_now = 12

    # calculate the difference between lollipops Jason had originally had lollipops he has now
    lollipops_given_to_denny = lollipops_jason_before - lollipops_jason_now
    return lollipops_given_to_denny


Q: Shawn has five toys. For Christmas, he got two toys each from his mom and dad. How many toys does he have now?

A:
def solution():
    shawn_toys_before = 5
    new_toys_from_each_parent = 2

    # first, calculate how many toys in total Shawn got for Christmas
    new_toys_total = 2 * new_toys_from_each_parent
    # next, calculate the sum of the number of toys Shawn originally had and the number of new toys he got
    shawn_toys_now = shawn_toys_before + new_toys_total
    return shawn_toys_now


Q: There were nine computers in the server room. Five more computers were installed each day, from monday to thursday. How many computers are now in the server room?

A:
def solution():
    computers_before = 9
    computers_installed_per_day = 5
    number_of_days = 4

    # first, calculate how many new computers were installed in total
    computers_installed_total = number_of_days * computers_installed_per_day
    # next, calculate sum of the number of computers originally in the room and the total number of new computers installed
    computers_now = computers_before + computers_installed_total
    return computers_now


Q: Michael had 58 golf balls. On tuesday, he lost 23 golf balls. On wednesday, he lost 2 more. How many golf balls did he have at the end of wednesday?

A:
def solution():
    golf_balls_before = 58
    golf_balls_lost_tuesday = 23
    golf_balls_lost_wednesday = 2

    # first, calculate the total number of golf balls Michael lost
    golf_balls_lost_total = golf_balls_lost_tuesday + golf_balls_lost_wednesday
    # next, calculate the difference between the original number of golf balls and the number of golf balls lost
    golf_balls_left = golf_balls_before - golf_balls_lost_total
    return golf_balls_left


Q: Olivia has $23. She bought five bagels for $3 each. How much money does she have left?

A:
def solution():
    dollars_before = 23
    bagels = 5
    price_per_bagel = 3

    # first, calculate the total cost of the bagels
    bagels_cost_total = bagels * price_per_bagel
    # next, calculate the difference between Olivia's original money and what she paid for the bagels
    dollars_left = dollars_before - bagels_cost_total
    return dollars_left


Q: <Question here>

A:
def solution():
\end{lstlisting}

\subsection*{Structured Code Prompt}
\begin{lstlisting}[caption={\textit{Structured code prompt format.}},label={lst:code-long}]
As an expert problem solver and Python developer, write Python functions returning numerical solutions to the following mathematical questions.

Q: There are 15 trees in the grove. Grove workers will plant trees in the grove today. After they are done, there will be 21 trees. How many trees did the grove workers plant today?

A:
def solution():
    # given:
    trees_now = 15  # number of trees currently in the grove
    trees_after_planting = 21  # number of trees in the grove after the workers are done planting today

    # to find: number of trees the grove workers will plant today

    # solution:
    # calculate the difference between trees after planting and trees originally
    trees_to_plant_today = trees_after_planting - trees_now
    return trees_to_plant_today


Q: If there are 3 cars in the parking lot and 2 more cars arrive, how many cars are in the parking lot?

A:
def solution():
    # given:
    cars_before = 3  # number of cars originally in the parking lot
    cars_arriving = 2  # number of cars arriving in the parking lot

    # to find: total number of cars in the parking lot

    # solution:
    # calculate the sum of cars originally in the parking lot and the arriving cars
    cars_after = cars_before + cars_arriving
    return cars_after


Q: Leah had 32 chocolates and her sister had 42. If they ate 35, how many pieces do they have left in total?

A:
def solution():
    # given:
    chocolates_leah_before = 32  # number of chocolates Leah originally had
    chocolates_sister_before = 42  # number of chocolates Leah's sister originally had
    chocolates_eaten = 35  # number of chocolates Leah and her sister ate

    # to find: total number of chocolates left

    # solution:
    # first, calculate how many chocolates the sisters originally had in total
    chocolates_total_before = chocolates_leah_before + chocolates_sister_before
    # next, subtract the total number of chocolates they ate
    chocolates_left = chocolates_total_before - chocolates_eaten
    return chocolates_left


Q: Jason had 20 lollipops. He gave Denny some lollipops. Now Jason has 12 lollipops. How many lollipops did Jason give to Denny?

A:
def solution():
    # given:
    lollipops_jason_before = 20  # number of lollipops Jason had originally
    lollipops_jason_now = 12  # number of lollipops Jason has left

    # to find: number of lollipops Jason gave to Denny

    # solution:
    # calculate the difference between lollipops Jason had originally had lollipops he has now
    lollipops_given_to_denny = lollipops_jason_before - lollipops_jason_now
    return lollipops_given_to_denny


Q: Shawn has five toys. For Christmas, he got two toys each from his mom and dad. How many toys does he have now?

A:
def solution():
    # given:
    shawn_toys_before = 5  # number of toys Shawn originally has
    new_toys_from_each_parent = 2  # number of toys his mom and his dad each gave him for Christmas

    # to find: total number of toys Shawn has now

    # solution:
    # first, calculate how many toys in total Shawn got for Christmas
    new_toys_total = 2 * new_toys_from_each_parent
    # next, calculate the sum of the number of toys Shawn originally had and the number of new toys he got
    shawn_toys_now = shawn_toys_before + new_toys_total
    return shawn_toys_now


Q: There were nine computers in the server room. Five more computers were installed each day, from monday to thursday. How many computers are now in the server room?

A:
def solution():
    # given:
    computers_before = 9  # number of computers originally in the server room
    computers_installed_per_day = 5  # number of new computers installed each day
    number_of_days = 4  # number of days from monday to thursday

    # to find: total number of computers in the server room now

    # solution:
    # first, calculate how many new computers were installed in total
    computers_installed_total = number_of_days * computers_installed_per_day
    # next, calculate sum of the number of computers originally in the room and the total number of new computers installed
    computers_now = computers_before + computers_installed_total
    return computers_now


Q: Michael had 58 golf balls. On tuesday, he lost 23 golf balls. On wednesday, he lost 2 more. How many golf balls did he have at the end of wednesday?

A:
def solution():
    # given:
    golf_balls_before = 58  # number of golf balls Michael originally had
    golf_balls_lost_tuesday = 23  # number of golf balls lost on tuesday
    golf_balls_lost_wednesday = 2  # number of golf balls lost on wednesday

    # to find: number of golf balls left at the end of wednesday

    # solution:
    # first, calculate the total number of golf balls Michael lost
    golf_balls_lost_total = golf_balls_lost_tuesday + golf_balls_lost_wednesday
    # next, calculate the difference between the original number of golf balls and the number of golf balls lost
    golf_balls_left = golf_balls_before - golf_balls_lost_total
    return golf_balls_left


Q: Olivia has $23. She bought five bagels for $3 each. How much money does she have left?

A:
def solution():
    # given:
    dollars_before = 23  # number of dollars Olivia had originally
    bagels = 5  # number of bagels Olivia bought
    price_per_bagel = 3  # number of dollars each bagel cost

    # to find: how many dollars Olivia has left?

    # solution:
    # first, calculate the total cost of the bagels
    bagels_cost_total = bagels * price_per_bagel
    # next, calculate the difference between Olivia's original money and what she paid for the bagels
    dollars_left = dollars_before - bagels_cost_total
    return dollars_left


Q: <Question here>

A:
def solution():
\end{lstlisting}

\newpage
\section{Supplementary Results} \label{apx:full-results}
In this section, we present supplementary figures and tables detailing the results of our variant effect and large-number effect testing experiments. 

\subsection{Ex1: Reproducing the GSM-Symbolic Benchmark}
Tables \ref{tab:GSM-results} and \ref{tab:GSM-stats} show mean accuracies, variant performance deltas, and results corresponding statistical significance testing following our reproduction of the GSM-Symbolic benchmark. The variant effect is statistically significant in 40\% of the tested models. 

\begin{table}[H]
\centering
\small
\begin{tabular}{lccccc}
\toprule
Model & $\overline{A}_{GSM-Base}$ & $\overline{A}_{GSM-Variants}$ & $\Delta_{var}$ & P value & Corrected P value \\
\midrule
gemma-2b & 19.0 & 12.0 & -6.96 & \textbf{0.030} & 0.443 \\
gemma-2b-it & 10.0 & 7.5 & -2.52 & 0.380 & 1.000 \\
gemma-7b & 48.0 & 47.7 & -0.30 & 0.954 & 1.000 \\
gemma-7b-it & 31.0 & 22.8 & -8.22 & \textbf{0.043} & 0.588 \\
gemma-2-2b & 24.0 & 20.9 & -3.12 & 0.321 $\ddagger$ & 1.000 \\
gemma-2-2b-it & 42.0 & 41.8 & -0.18 & 0.993 & 1.000 \\
gemma-2-9b & 69.0 & 61.1 & -7.88 & \textbf{0.024} & 0.432 \\
gemma-2-9b-it & 86.0 & 82.6 & -3.40 & 0.227 & 1.000 \\
gemma-2-27b-it & 90.0 & 87.6 & -2.40 & 0.378 & 1.000 \\
phi-2 & 60.0 & 39.7 & -20.26 & \textbf{<0.001} & \textbf{<0.001} \\
Phi-3-mini-128k-instruct & 80.0 & 78.7 & -1.30 & 0.710 & 1.000 \\
Phi-3-medium-128k-instruct & 91.0 & 88.2 & -2.78 & 0.304 & 1.000 \\
Phi-3.5-mini-instruct & 87.0 & 79.4 & -7.58 & \textbf{0.013} & 0.250 \\
Mistral-7B-v0.1 & 38.0 & 33.0 & -5.00 & 0.190 & 1.000 \\
Mistral-7B-Instruct-v0.1 & 36.0 & 27.5 & -8.54 & \textbf{0.025} & 0.432 \\
Mistral-7B-v0.3 & 33.0 & 35.5 & 2.50 & 0.511 & 1.000 \\
Mistral-7B-Instruct-v0.3 & 52.0 & 45.3 & -6.68 & 0.072 & 0.864 \\
Mathstral-7B-v0.1 & 82.0 & 75.1 & -6.88 & \textbf{0.042} & 0.588 \\
Meta-Llama-3-8B & 55.0 & 47.4 & -7.56 & \textbf{0.026} & 0.432 \\
Meta-Llama-3-8B-Instruct & 77.0 & 69.9 & -7.08 & 0.078 $\ddagger$ & 0.864 \\
\bottomrule
\end{tabular}
\caption{Results of the variant effect testing (Ex1) for the \textit{GSM prompt}, as returned by GLMM 1. \pbdesc\ \ddagdesc}
\label{tab:GSM-results}
\end{table}

\begin{table}[H]
\centering
\small
\begin{tabular}{lcccc}
\toprule
Model & Odds ratio & 95\% CI & Rand. eff. variance $\pm$ std. dev. & N estimates \\
\midrule
gemma-2b & 0.42 & [0.22, 0.91] & 6.53 $\pm$ 2.56 & 1963 \\
gemma-2b-it & 0.63 & [0.26, 2.10] & 5.41 $\pm$ 2.32 & 1996 \\
gemma-7b & 0.98 & [0.53, 1.71] & 6.49 $\pm$ 2.55 & 2000 \\
gemma-7b-it & 0.50 & [0.25, 0.97] & 5.00 $\pm$ 2.24 & 1955 \\
gemma-2-2b & 0.73 & [0.39, 1.41] & 7.22 $\pm$ 2.69 & 1207 \\
gemma-2-2b-it & 0.99 & [0.52, 1.86] & 9.68 $\pm$ 3.11 & 2000 \\
gemma-2-9b & 0.43 & [0.17, 0.92] & 10.20 $\pm$ 3.19 & 2000 \\
gemma-2-9b-it & 0.60 & [0.18, 1.31] & 11.05 $\pm$ 3.32 & 1984 \\
gemma-2-27b-it & 0.58 & [0.14, 1.81] & 17.39 $\pm$ 4.17 & 1835 \\
phi-2 & 0.23 & [0.13, 0.39] & 6.14 $\pm$ 2.48 & 2000 \\
Phi-3-mini-128k-instruct & 0.85 & [0.38, 1.76] & 8.95 $\pm$ 2.99 & 1992 \\
Phi-3-medium-128k-instruct & 0.56 & [0.13, 1.56] & 9.50 $\pm$ 3.08 & 1960 \\
Phi-3.5-mini-instruct & 0.33 & [0.11, 0.79] & 8.49 $\pm$ 2.91 & 1974 \\
Mistral-7B-v0.1 & 0.67 & [0.37, 1.18] & 6.80 $\pm$ 2.61 & 1993 \\
Mistral-7B-Instruct-v0.1 & 0.50 & [0.28, 0.91] & 5.06 $\pm$ 2.25 & 1998 \\
Mistral-7B-v0.3 & 1.25 & [0.64, 2.43] & 7.08 $\pm$ 2.66 & 1998 \\
Mistral-7B-Instruct-v0.3 & 0.60 & [0.34, 1.04] & 6.98 $\pm$ 2.64 & 2000 \\
Mathstral-7B-v0.1 & 0.47 & [0.19, 0.96] & 6.46 $\pm$ 2.54 & 1999 \\
Meta-Llama-3-8B & 0.49 & [0.25, 0.91] & 11.25 $\pm$ 3.35 & 2000 \\
Meta-Llama-3-8B-Instruct & 0.48 & [0.21, 1.07] & 9.95 $\pm$ 3.16 & 1998 \\
\bottomrule
\end{tabular}
\caption{Additional statistics for results of \textit{GSM prompt}. \textit{N estimates} indicates the number of cleanly converged bootstrap iterations, over which the statistics are calculated.}
\label{tab:GSM-stats}
\end{table}

\newpage

Tables \ref{tab:mirzadeh-comparison} and \ref{tab:mirzadeh-comparison-deltas} include comparison of our results with those reported by \citet{mirzadeh2025gsm}. The results differ up to 22 percentage points (pp) on \textit{GSM-Base}, 27.16 percentage points on \textit{GSM-Variants}, and 24.1 percentage points in variant performance deltas. However, no systematic shift can be seen; the differences of variant performance deltas (i.e. \textit{delta of deltas}, ours vs Mirzadeh) has a mean of -0.54 percentage points, median of -0.8, and standard deviation of 8.34. There are multiple factors possibly contributing to the observed differences. A primary culprit is likely model versioning --- the exact model versions on Hugging Face might have changed since \citeauthor{mirzadeh2025gsm}'s study was published; the authors did not mention the exact checkpoints or revision hashes and we defaulted to the most recent versions available. Model sensitivity to minor prompt formatting and decoding could also be a source of differences if despite trying to reproduce the study as closely as possible, we missed fine details such as trailing white spacing in the prompts.

\begin{table}[H]
\centering
\small
\begin{tabular}{lcccc}
\toprule
 & GSM-Base / M & GSM-Variants / M & GSM-Base / O & GSM-Variants / O \\
\midrule
gemma-2b & 11.0 & 8.2 (± 2.21) & 19.0 & 12.0 (± 2.67) \\
gemma-2b-it & 11.0 & 8.2 (± 2.21) & 10.0 & 7.5 (± 2.09) \\
gemma-7b & 50.0 & 25.6 (± 3.25) & 48.0 & 47.7 (± 3.49) \\
gemma-7b-it & 33.0 & 25.6 (± 3.25) & 31.0 & 22.8 (± 2.74) \\
gemma-2-2b & 46.0 & 40.1 (± 3.04) & 24.0 & 20.9 (± 3.13) \\
gemma-2-2b-it & 46.0 & 40.1 (± 3.04) & 42.0 & 41.8 (± 3.11) \\
gemma-2-9b & 87.0 & 79.1 (± 2.99) & 69.0 & 61.1 (± 3.36) \\
gemma-2-9b-it & 87.0 & 79.1 (± 2.99) & 86.0 & 82.6 (± 3.00) \\
gemma-2-27b-it & 92.0 & 88.3 (± 2.56) & 90.0 & 87.6 (± 1.92) \\
phi-2 & 53.0 & 41.4 (± 3.56) & 60.0 & 39.7 (± 3.39) \\
Phi-3-mini-128k-instruct & 85.0 & 80.7 (± 2.94) & 80.0 & 78.7 (± 2.88) \\
Phi-3-medium-128k-instruct & 89.0 & 82.5 (± 2.86) & 91.0 & 88.2 (± 2.57) \\
Phi-3.5-mini-instruct & 88.0 & 82.1 (± 3.38) & 87.0 & 79.4 (± 2.43) \\
Mistral-7B-v0.1 & 48.0 & 41.1 (± 3.36) & 38.0 & 33.0 (± 3.28) \\
Mistral-7B-Instruct-v0.1 & 42.0 & 30.5 (± 3.47) & 36.0 & 27.5 (± 3.92) \\
Mistral-7B-v0.3 & 44.0 & 40.0 (± 4.43) & 33.0 & 35.5 (± 3.04) \\
Mistral-7B-Instruct-v0.3 & 56.0 & 50.0 (± 3.49) & 52.0 & 45.3 (± 3.68) \\
Mathstral-7B-v0.1 & 80.0 & 74.0 (± 3.49) & 82.0 & 75.1 (± 3.17) \\
Meta-Llama-3-8B & 61.0 & 74.6 (± 2.94) & 55.0 & 47.4 (± 3.20) \\
Meta-Llama-3-8B-Instruct & 74.0 & 74.6 (± 2.94) & 77.0 & 69.9 (± 3.28) \\
\bottomrule
\end{tabular}
\caption{Comparison of results reported in the GSM-Symbolic study \citeauthor{mirzadeh2025gsm} (marked with '/ M') and our reproduction of the benchmark (marked with '/ O').}
\label{tab:mirzadeh-comparison}
\end{table}

\begin{table}[H]
\centering
\small
\begin{tabular}{lcc}
\toprule
 & $\Delta_{var}$ / Mirzadeh & $\Delta_{var}$ / Ours \\
\midrule
gemma-2b & 2.8 & 7.0 \\
gemma-2b-it & 2.8 & 2.5 \\
gemma-7b & 24.4 & 0.3 \\
gemma-7b-it & 7.4 & 8.2 \\
gemma-2-2b & 5.9 & 3.1 \\
gemma-2-2b-it & 5.9 & 0.2 \\
gemma-2-9b & 7.9 & 7.9 \\
gemma-2-9b-it & 7.9 & 3.4 \\
gemma-2-27b-it & 3.7 & 2.4 \\
phi-2 & 11.6 & 20.3 \\
Phi-3-mini-128k-instruct & 4.3 & 1.3 \\
Phi-3-medium-128k-instruct & 6.5 & 2.8 \\
Phi-3.5-mini-instruct & 5.9 & 7.6 \\
Mistral-7B-v0.1 & 6.9 & 5.0 \\
Mistral-7B-Instruct-v0.1 & 11.5 & 8.5 \\
Mistral-7B-v0.3 & 4.0 & -2.5 \\
Mistral-7B-Instruct-v0.3 & 6.0 & 6.7 \\
Mathstral-7B-v0.1 & 6.0 & 6.9 \\
Meta-Llama-3-8B & -13.6 & 7.6 \\
Meta-Llama-3-8B-Instruct & -0.6 & 7.1 \\
\bottomrule
\end{tabular}
\caption{Comparison of variant performance deltas (changes in accuracy on \textit{GSM-Variants} vs \textit{GSM-Base} data) reported in the GSM-Symbolic study \citep{mirzadeh2025gsm} and in our reproduction of the benchmark.}
\label{tab:mirzadeh-comparison-deltas}
\end{table}

\newpage
\subsection{Alternative Prompt Formats}
\label{apx:prompt-formats-results}
This section presents supplementary results for variant effect testing when using the four alternative prompt formats (see Section~\ref{sec:prompt-formats}).

\subsubsection{Ex2.A: \textit{Simple NL Prompt}}

\begin{figure}[H]
  \centering
  \includegraphics[width=\linewidth]{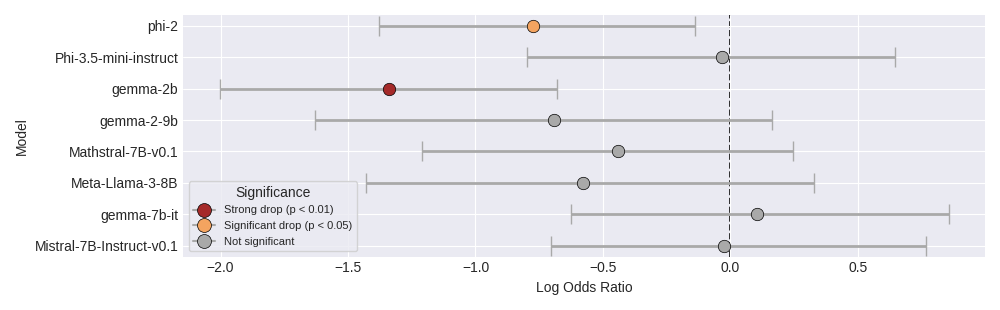}
  \includegraphics[width=\linewidth]{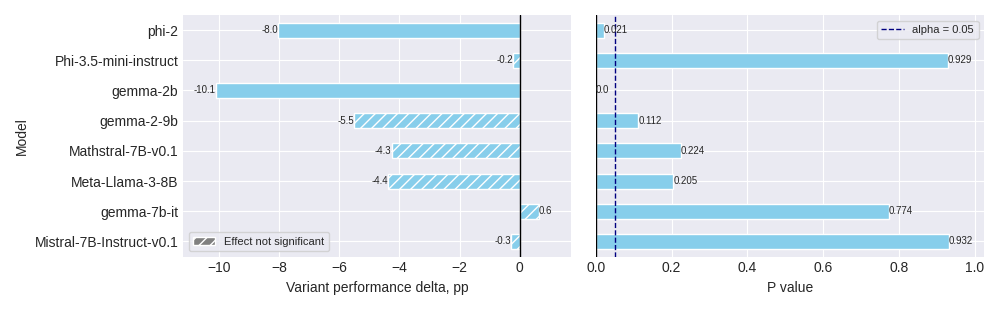}
  \caption{Values and significance of variant performance deltas for the \textit{simple NL prompt}. \plotdesc}
  \label{fig:nonformal}
\end{figure}

Figure~\ref{fig:nonformal}, as well as Tables~\ref{tab:nonformal-results} and \ref{tab:nonformal-stats}, present results of Ex2.A, i.e. variant effect testing on the \textit{simple NL prompt} (see Section~\ref{sec:prompt-formats} for prompt formats definitions). This adaptation of the original prompt format removes significance of variant performance delta in all but two of the models: phi-2 and gemma-2b.

\begin{table}[H]
\centering
\small  
\begin{tabular}{lccccc}
\toprule
Model & $\overline{A}_{GSM-Base}$ & $\overline{A}_{GSM-Variants}$ & $\Delta_{var}$ & P value & Corrected P value \\
\midrule
phi-2 & 45.0 & 37.0 & -8.04 & \textbf{0.021} & 0.147 \\
Phi-3.5-mini-instruct & 73.0 & 72.8 & -0.22 & 0.929 & 1.000 \\
gemma-2b & 21.0 & 10.9 & -10.12 & \textbf{< 0.001} & \textbf{< 0.001} \\
gemma-2-9b & 67.0 & 61.5 & -5.52 & 0.112 & 0.673 \\
Mathstral-7B-v0.1 & 73.0 & 68.7 & -4.26 & 0.224 & 1.000 \\
Meta-Llama-3-8B & 21.0 & 16.6 & -4.38 & 0.205 & 1.000 \\
gemma-7b-it & 24.0 & 24.6 & 0.64 & 0.774 & 1.000 \\
Mistral-7B-Instruct-v0.1 & 26.0 & 25.7 & -0.30 & 0.932 & 1.000 \\
\bottomrule
\end{tabular}
\caption{Results of variant effect testing obtained with the \textit{simple NL prompt}, as returned by GLMM 1. \pbdesc}
\label{tab:nonformal-results}
\end{table}

\begin{table}[H]
\centering
\small
\begin{tabular}{lcccc}
\toprule
Model & Odds ratio & 95\% CI & Rand. eff. variance $\pm$ std. dev. & N estimates \\
\midrule
phi-2 & 0.46 & [0.25, 0.87] & 13.44 $\pm$ 3.67 & 1998 \\
Phi-3.5-mini-instruct & 0.97 & [0.45, 1.91] & 12.20 $\pm$ 3.49 & 1995 \\
gemma-2b & 0.26 & [0.13, 0.51] & 6.98 $\pm$ 2.64 & 1992 \\
gemma-2-9b & 0.50 & [0.20, 1.18] & 15.73 $\pm$ 3.97 & 1998 \\
Mathstral-7B-v0.1 & 0.64 & [0.30, 1.28] & 9.72 $\pm$ 3.12 & 2000 \\
Meta-Llama-3-8B & 0.56 & [0.24, 1.39] & 12.15 $\pm$ 3.49 & 1973 \\
gemma-7b-it & 1.11 & [0.53, 2.35] & 14.78 $\pm$ 3.85 & 1905 \\
Mistral-7B-Instruct-v0.1 & 0.98 & [0.50, 2.16] & 10.68 $\pm$ 3.27 & 1528 \\
\bottomrule
\end{tabular}
\caption{Additional statistics for results of \textit{simple NL prompt}.}
\label{tab:nonformal-stats}
\end{table}

\subsubsection{Ex2.B: \textit{Structured NL Prompt}}
\begin{figure}[H]
  \centering
  \includegraphics[width=\linewidth]{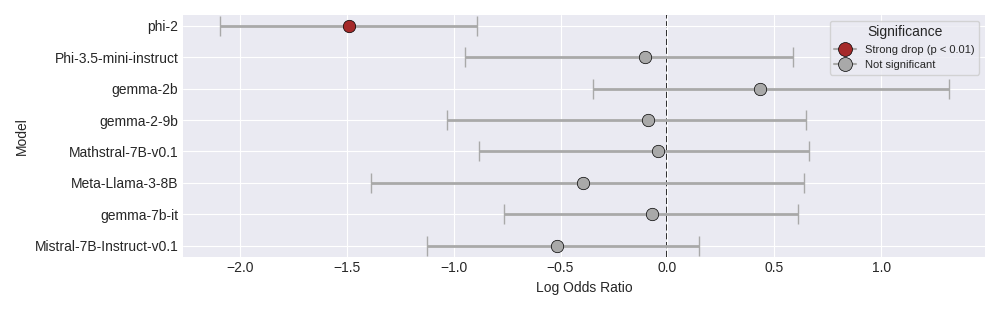}
  \includegraphics[width=\linewidth]{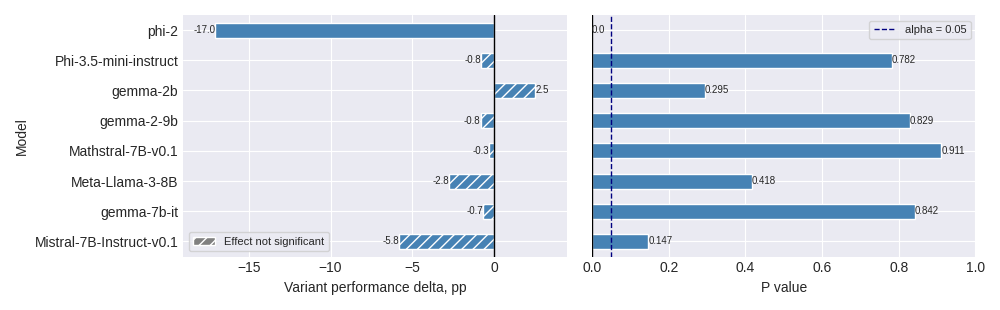}
  \caption{Values and significance of variant performance deltas for the structured natural language prompt. \plotdesc\ }
  \label{fig:formal}
\end{figure}

Figure~\ref{fig:formal}, supported by Tables~\ref{tab:nonformal-results} and Table~\ref{tab:nonformal-stats}, depicts the results for experiment Ex2.B: \textit{structured NL prompt} (see Section~\ref{sec:prompt-formats} for prompt formats definitions). The variant effect remains statistically significant for phi-2.

\begin{table}[H]
\centering
\small
\begin{tabular}{lccccc}
\toprule
Model & $\overline{A}_{GSM-Base}$ & $\overline{A}_{GSM-Variants}$ & $\Delta_{var}$ & P value & Corrected P value \\
\midrule
phi-2 & 63.0 & 46.0 & -17.04 & \textbf{< 0.001} & \textbf{< 0.001} \\
Phi-3.5-mini-instruct & 77.0 & 76.2 & -0.80 & 0.782 & 1.000 \\
gemma-2b & 12.0 & 14.5 & 2.50 & 0.295 & 1.000 \\
gemma-2-9b & 72.0 & 71.2 & -0.84 & 0.829 & 1.000 \\
Mathstral-7B-v0.1 & 78.0 & 77.7 & -0.34 & 0.911 & 1.000 \\
Meta-Llama-3-8B & 16.0 & 13.2 & -2.78 & 0.418 & 1.000 \\
gemma-7b-it & 29.0 & 28.3 & -0.68 & 0.842 & 1.000 \\
Mistral-7B-Instruct-v0.1 & 35.0 & 29.2 & -5.82 & 0.147 & 1.000 \\
\bottomrule
\end{tabular}
\caption{Variant effect testing results obtained with the \textit{structured NL prompt} (see Section~\ref{sec:prompt-formats} for prompt formats definitions), as returned by GLMM 1. \pbdesc}
\label{tab:formal-results}
\end{table}

\begin{table}[H]
\centering
\small
\begin{tabular}{lcccc}
\toprule
Model & Odds ratio & 95\% CI & Rand. eff. variance $\pm$ std. dev. & N estimates \\
\midrule
phi-2 & 0.23 & [0.12, 0.41] & 11.83 $\pm$ 3.44 & 2000 \\
Phi-3.5-mini-instruct & 0.90 & [0.39, 1.80] & 9.82 $\pm$ 3.13 & 1995 \\
gemma-2b & 1.54 & [0.71, 3.72] & 7.48 $\pm$ 2.73 & 1961 \\
gemma-2-9b & 0.91 & [0.36, 1.91] & 15.07 $\pm$ 3.88 & 1989 \\
Mathstral-7B-v0.1 & 0.96 & [0.41, 1.93] & 7.46 $\pm$ 2.73 & 1993 \\
Meta-Llama-3-8B & 0.67 & [0.25, 1.89] & 10.34 $\pm$ 3.22 & 1954 \\
gemma-7b-it & 0.93 & [0.47, 1.84] & 10.47 $\pm$ 3.24 & 1715 \\
Mistral-7B-Instruct-v0.1 & 0.60 & [0.32, 1.16] & 8.77 $\pm$ 2.96 & 1949 \\
\bottomrule
\end{tabular}
\caption{Additional statistics for the variant effect testing with the \textit{structured NL prompt}, as returned by GLMM 1.}
\label{tab:formal-stats}
\end{table}

\subsubsection{Ex2.C: \textit{Simple Code Prompt}}
\begin{figure}[H]
\centering
  \includegraphics[width=\linewidth]{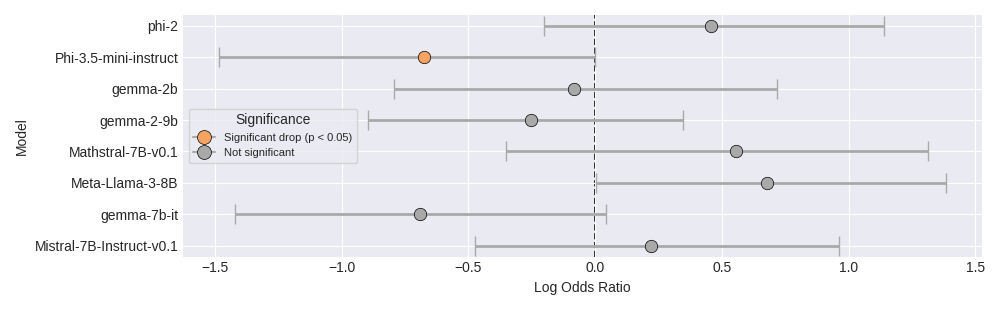}
  \includegraphics[width=\linewidth]{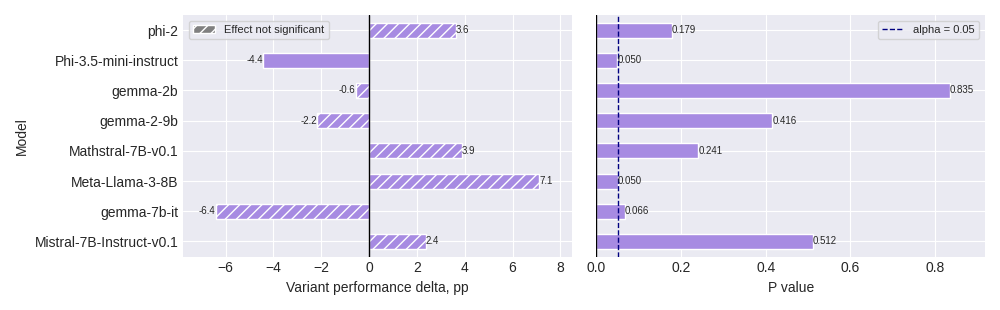}
  \caption{Values and significance of variant performance deltas for the \textit{simple code prompt}. \plotdesc}
  \label{fig:code-short}
\end{figure}

Variant performance delta results for the \textit{simple code prompt} (see Section~\ref{sec:prompt-formats}) are shown in Figure~\ref{fig:code-short} and Tables \ref{tab:code-short-results} and \ref{tab:code-short-stats}. The variant effect is significant for Phi-3.5-mini-instruct.

\begin{table}[H]
\centering
\small
\begin{tabular}{lccccc}
\toprule
Model & $\overline{A}_{GSM-Base}$ & $\overline{A}_{GSM-Variants}$ & $\Delta_{var}$ & P value & Corrected P value \\
\midrule
phi-2 & 63.0 & 66.6 & 3.64 & 0.179 & 0.894 \\
Phi-3.5-mini-instruct & 81.0 & 76.6 & -4.42 & \textbf{0.050} $\ddagger$ & 0.396 \\
gemma-2b & 14.0 & 13.4 & -0.56 & 0.835 & 1.000 \\
gemma-2-9b & 68.0 & 65.8 & -2.18 & 0.416 & 1.000 \\
Mathstral-7B-v0.1 & 75.0 & 78.9 & 3.88 & 0.241 & 0.963 \\
Meta-Llama-3-8B & 48.0 & 55.1 & 7.12 & 0.050 $\ddagger$ & 0.396 \\
gemma-7b-it & 47.0 & 40.6 & -6.42 & 0.066 $\ddagger$ & 0.399 \\
Mistral-7B-Instruct-v0.1 & 41.0 & 43.4 & 2.36 & 0.512 & 1.000 \\
\bottomrule
\end{tabular}
\caption{Variant effect testing results obtained with the \textit{simple code prompt}, as returned by GLMM 1. \pbdesc\ \ddagdesc}
\label{tab:code-short-results}
\end{table}

\begin{table}[H]
\centering
\small
\begin{tabular}{lcccc}
\toprule
Model & Odds ratio & 95\% CI & Rand. eff. variance $\pm$ std. dev. & N estimates \\
\midrule
phi-2 & 1.58 & [0.82, 3.12] & 18.55 $\pm$ 4.31 & 1990 \\
Phi-3.5-mini-instruct & 0.51 & [0.23, 1.00] & 19.85 $\pm$ 4.46 & 1898 \\
gemma-2b & 0.92 & [0.45, 2.05] & 6.28 $\pm$ 2.51 & 1864 \\
gemma-2-9b & 0.78 & [0.41, 1.41] & 15.63 $\pm$ 3.95 & 1995 \\
Mathstral-7B-v0.1 & 1.74 & [0.70, 3.72] & 19.00 $\pm$ 4.36 & 1903 \\
Meta-Llama-3-8B & 1.97 & [1.00, 3.99] & 10.66 $\pm$ 3.26 & 2000 \\
gemma-7b-it & 0.50 & [0.24, 1.04] & 14.41 $\pm$ 3.80 & 1926 \\
Mistral-7B-Instruct-v0.1 & 1.25 & [0.62, 2.61] & 11.76 $\pm$ 3.43 & 2000 \\
\bottomrule
\end{tabular}
\caption{Additional statistics for the variant effect testing with the \textit{simple code prompt}, as returned by GLMM 1.}
\label{tab:code-short-stats}
\end{table}

\subsubsection{Ex2.D: \textit{Structured Code Prompt}}
\begin{figure}[H]
\centering
  \includegraphics[width=\linewidth]{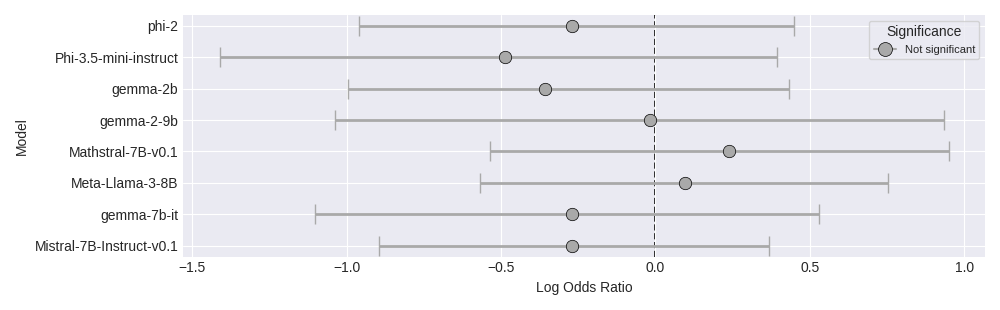}
  \includegraphics[width=\linewidth]{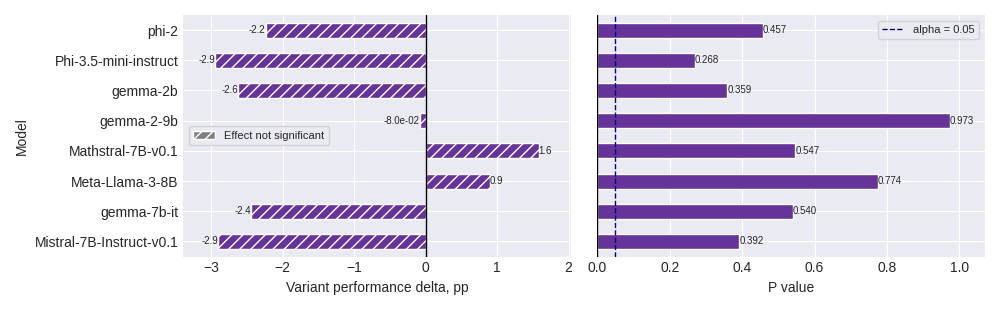}
  \caption{Values and significance of variant performance deltas for the \textit{structured code prompt}. \plotdesc}
  \label{fig:code-long}
\end{figure}

Figure~\ref{fig:code-long} and Tables \ref{tab:code-long-results} and \ref{tab:code-long-stats} present variant effect testing results for the \textit{structured code prompt} (see Section~\ref{sec:prompt-formats}). Variant effect is not significant for any of the tested LLMs.

\begin{table}[H]
\centering
\small
\begin{tabular}{lccccc}
\toprule
Model & $\overline{A}_{GSM-Base}$ & $\overline{A}_{GSM-Variants}$ & $\Delta_{var}$ & P value & Corrected P value \\
\midrule
phi-2 & 66.0 & 63.8 & -2.24 & 0.457 & 1.000 \\
Phi-3.5-mini-instruct & 83.0 & 80.1 & -2.94 & 0.268 & 1.000 \\
gemma-2b & 16.0 & 13.4 & -2.62 & 0.359 & 1.000 \\
gemma-2-9b & 68.0 & 67.9 & -0.08 & 0.973 & 1.000 \\
Mathstral-7B-v0.1 & 78.0 & 79.6 & 1.58 & 0.547 & 1.000 \\
Meta-Llama-3-8B & 49.0 & 49.9 & 0.90 & 0.774 & 1.000 \\
gemma-7b-it & 43.0 & 40.6 & -2.44 & 0.540 & 1.000 \\
Mistral-7B-Instruct-v0.1 & 47.0 & 44.1 & -2.90 & 0.392 & 1.000 \\
\bottomrule
\end{tabular}
\caption{Variant effect testing results obtained with the \textit{structured code prompt}, as returned by GLMM 1.}
\label{tab:code-long-results}
\end{table}

\begin{table}[H]
\centering
\small
\begin{tabular}{lcccc}
\toprule
Model & Odds ratio & 95\% CI & Rand. eff. variance $\pm$ std. dev. & N estimates \\
\midrule
phi-2 & 0.76 & [0.38, 1.57] & 18.91 $\pm$ 4.35 & 1980 \\
Phi-3.5-mini-instruct & 0.61 & [0.24, 1.48] & 25.71 $\pm$ 5.07 & 1863 \\
gemma-2b & 0.70 & [0.37, 1.54] & 6.30 $\pm$ 2.51 & 1872 \\
gemma-2-9b & 0.98 & [0.35, 2.55] & 16.70 $\pm$ 4.09 & 1987 \\
Mathstral-7B-v0.1 & 1.27 & [0.58, 2.58] & 19.08 $\pm$ 4.37 & 1880 \\
Meta-Llama-3-8B & 1.10 & [0.57, 2.13] & 15.81 $\pm$ 3.98 & 2000 \\
gemma-7b-it & 0.76 & [0.33, 1.70] & 16.29 $\pm$ 4.04 & 1844 \\
Mistral-7B-Instruct-v0.1 & 0.76 & [0.41, 1.45] & 10.82 $\pm$ 3.29 & 2000 \\
\bottomrule
\end{tabular}
\caption{Additional statistics for the variant effect testing with the \textit{structured code prompt}, as returned by GLMM~1.}
\label{tab:code-long-stats}
\end{table}

\newpage
\subsection{The Large-Number Effect}\label{apx:number-effect}
\begin{figure}[H]
  \centering
  \includegraphics[width=0.9\linewidth]{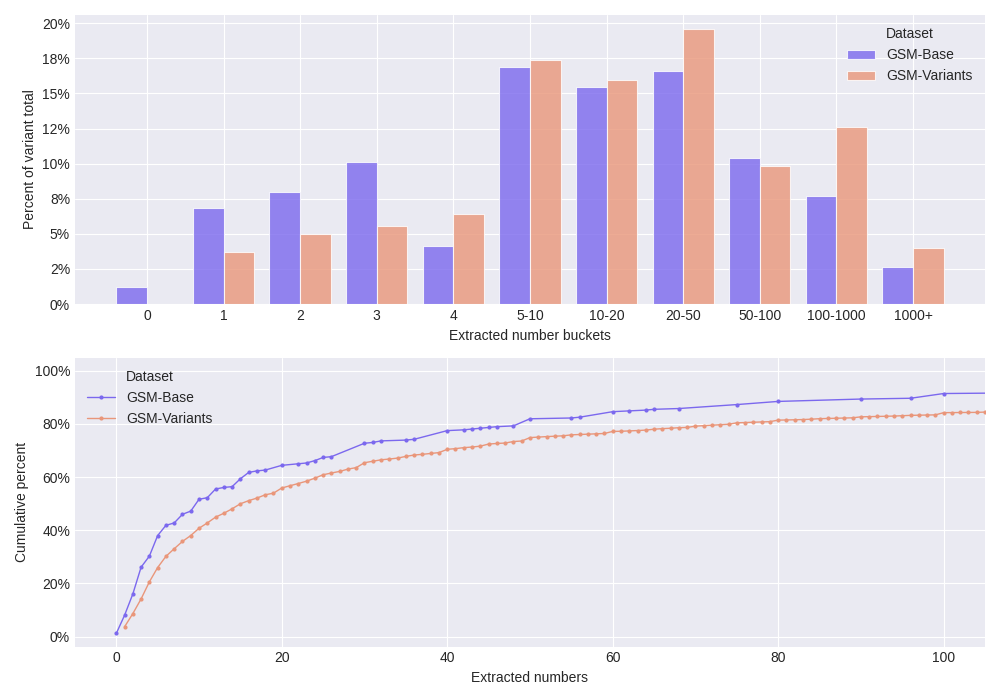}
  \caption{Distribution of integers extracted from problem texts in \textit{GSM-Base} and \textit{GSM-Variants} datasets. Top: counts of integers grouped into variable-width bins, normalised by the total of integers extracted from each dataset's questions. Bottom: normalised cumulative counts of extracted numbers, zoomed for clearer presentation of the distribution shift.}
  \label{fig:numbers}
\end{figure}

Figure~\ref{fig:numbers} visualises the difference in distribution of integers occurring in questions from the \textit{GSM-Base} and \textit{GSM-Variants} datasets. In the latter, the distribution is shifted towards larger numbers with K-S statistic of 12 percentage points and high statistical significance ($p < 0.001$), as reported in Section~\ref{sec:number-effect}.

\newpage
Tables~\ref{tab:number_effect_odds} and \ref{tab:corrected_delta_symb_odds} present the results of the large-number effect significance testing for the 8 evaluated LLMs.

\begin{table}[H]
\small
\centering
\begin{tabular}{lccccc}
\toprule
prompt & GSM & NL-simple & NL-structured & code-simple & code-structured \\
model &  &  &  &  &  \\
\midrule
phi-2 & \makecell{0.76 \\ \textbf{0.003} / \textbf{0.012}} & \makecell{0.72 \\ \textbf{0.008} / \textbf{0.024}} & \makecell{0.66 \\ \textbf{<0.001} / \textbf{<0.001}} & \makecell{0.90 \\ 0.480 / 0.961} & \makecell{1.00 \\ 0.981 / 0.981} \\ \addlinespace
Phi-3.5-mini-instruct & \makecell{0.74 \\ \textbf{0.001} / \textbf{0.005}} & \makecell{0.77 \\ \textbf{0.001} / \textbf{0.005}} & \makecell{0.81 \\ \textbf{0.008} / \textbf{0.023}} & \makecell{1.03 \\ 0.828 / 0.828} & \makecell{0.88 \\ 0.232 / 0.463} \\ \addlinespace
gemma-2b & \makecell{0.85 \\ 0.065 $\ddagger$ / 0.194} & \makecell{0.80 \\ \textbf{0.033} / 0.132} & \makecell{0.78 \\ \textbf{0.004} / \textbf{0.020}} & \makecell{0.98 \\ 0.778 / 1.000} & \makecell{0.97 \\ 0.731 / 1.000} \\ \addlinespace
gemma-2-9b & \makecell{1.04 \\ 0.722 / 1.000} & \makecell{0.74 \\ \textbf{0.028} / 0.138} & \makecell{0.90 \\ 0.365 / 1.000} & \makecell{1.15 \\ 0.270 / 1.000} & \makecell{0.94 \\ 0.538 / 1.000} \\ \addlinespace
Mathstral-7B-v0.1 & \makecell{0.90 \\ 0.260 / 0.779} & \makecell{0.79 \\ \textbf{0.031} / 0.124} & \makecell{0.82 \\ \textbf{0.025} / 0.124} & \makecell{0.88 \\ 0.266 / 0.779} & \makecell{0.94 \\ 0.645 / 0.779} \\ \addlinespace
Meta-Llama-3-8B & \makecell{1.09 \\ 0.463 / 1.000} & \makecell{0.80 \\ 0.066 $\ddagger$ / 0.328} & \makecell{1.06 \\ 0.644 / 1.000} & \makecell{1.12 \\ 0.305 / 1.000} & \makecell{0.96 \\ 0.683 / 1.000} \\ \addlinespace
gemma-7b-it & \makecell{0.88 \\ \textbf{0.045} / 0.223} & \makecell{0.87 \\ 0.136 / 0.408} & \makecell{0.85 \\ 0.056 $\ddagger$ / 0.225} & \makecell{0.86 \\ 0.215 / 0.430} & \makecell{0.87 \\ 0.308 / 0.430} \\ \addlinespace
Mistral-7B-Instruct-v0.1 & \makecell{0.77 \\ \textbf{0.002} / \textbf{0.010}} & \makecell{0.85 \\ \textbf{0.034} / 0.101} & \makecell{0.79 \\ \textbf{0.005} / \textbf{0.022}} & \makecell{1.03 \\ 0.726 / 1.000} & \makecell{1.04 \\ 0.695 / 1.000} \\ \addlinespace
\bottomrule
\end{tabular}
\caption{Odds ratios and significance \textbf{of the large-number effect} across models and prompt formats, as returned by GLMM 2. Cells formatted as: (top row) odds ratio; (bottom row) raw P~value / Holm-Bonferroni-corrected (column-wise) P~value. \pbdesc\ \ddagdesc}
\label{tab:number_effect_odds}
\end{table}

\newpage

\begin{table}[H]
\small
\centering
\begin{tabular}{lccccc}
\toprule
prompt & GSM & NL-simple & NL-structured & code-simple & code-structured \\
model &  &  &  &  &  \\
\midrule
phi-2 & \makecell{0.31 \\ \textbf{<0.001} / \textbf{<0.001}} & \makecell{0.68 \\ 0.224 / 0.448} & \makecell{0.35 \\ \textbf{<0.001} / \textbf{<0.001}} & \makecell{1.70 \\ 0.141 / 0.423} & \makecell{0.76 \\ 0.480 / 0.480} \\ \addlinespace
Phi-3.5-mini-instruct & \makecell{0.59 \\ 0.239 / 0.954} & \makecell{1.41 \\ 0.340 / 1.000} & \makecell{1.16 \\ 0.683 $\ddagger$ / 1.000} & \makecell{0.49 \\ \textbf{0.045} $\ddagger$ / 0.223} & \makecell{0.72 \\ 0.483 / 1.000} \\ \addlinespace
gemma-2b & \makecell{0.49 \\ 0.081 $\ddagger$ / 0.242} & \makecell{0.33 \\ \textbf{0.003} / \textbf{0.015}} & \makecell{2.10 \\ 0.060 / 0.242} & \makecell{0.95 \\ 0.909 / 0.909} & \makecell{0.72 \\ 0.413 / 0.825} \\ \addlinespace
gemma-2-9b & \makecell{0.40 \\ \textbf{0.013} / 0.065} & \makecell{0.72 \\ 0.437 / 1.000} & \makecell{1.03 \\ 0.967 / 1.000} & \makecell{0.65 \\ 0.216 / 0.864} & \makecell{1.01 \\ 0.984 / 1.000} \\ \addlinespace
Mathstral-7B-v0.1 & \makecell{0.49 \\ \textbf{0.046} $\ddagger$ / 0.231} & \makecell{0.86 \\ 0.673 / 1.000} & \makecell{1.28 \\ 0.490 / 1.000} & \makecell{1.90 \\ 0.184 / 0.735} & \makecell{1.22 \\ 0.637 / 1.000} \\ \addlinespace
Meta-Llama-3-8B & \makecell{0.45 \\ \textbf{0.013} / 0.065} & \makecell{0.66 \\ 0.335 / 0.997} & \makecell{0.63 \\ 0.332 / 0.997} & \makecell{1.76 \\ 0.093 / 0.372} & \makecell{1.14 \\ 0.699 / 0.997} \\ \addlinespace
gemma-7b-it & \makecell{0.58 \\ 0.133 / 0.663} & \makecell{1.29 \\ 0.506 / 1.000} & \makecell{1.12 \\ 0.742 $\ddagger$ / 1.000} & \makecell{0.59 \\ 0.168 / 0.671} & \makecell{0.90 \\ 0.815 / 1.000} \\ \addlinespace
Mistral-7B-Instruct-v0.1 & \makecell{0.66 \\ 0.207 / 1.000} & \makecell{1.14 \\ 0.702 $\ddagger$ / 1.000} & \makecell{0.79 \\ 0.486 / 1.000} & \makecell{1.21 \\ 0.626 / 1.000} & \makecell{0.73 \\ 0.356 / 1.000} \\ \addlinespace
\bottomrule
\end{tabular}
\caption{Odds ratios and significance \textbf{of the large-number-effect-corrected variant effect} across models and prompt formats, as returned by GLMM 2. Cells formatted as: (top row) odds ratio; (bottom row) raw P~value / Holm-Bonferroni-corrected (column-wise) P~value. \pbdesc\ \ddagdesc}
\label{tab:corrected_delta_symb_odds}
\end{table}

\newpage

\section{Wald-based Significance Testing}\label{apx:degenerate-fit}
As mentioned in text, our initial approach was to use Wald method, i.e. only performing a single GLMM fit per dataset, assuming normal distribution of the responses. However, for several model-prompt format combinations, we observed encountered convergence issues in either instance of the model (variant effect / large-number effect estimation), which points to limited reliability of the approach for the data involved in this work.

\subsection{Wald-to-Bootstrap Mismatch}
In majority of cases, Wald-derived statistical significance matches the results obtained through bootstrapping, as demonstrated in results tables throughout the appendices. In Table~\ref{tab:wald-mismatch} below we list the cases where the two methods yielded opposing results.

\begin{table}[H]
\small
\centering
\begin{tabular}{cccccccc}
\toprule
Test & Effect & Prompt & Model & Bootstrap & Wald & Convergent \\
\midrule
GLMM 2 & Variant & GSM & Mathstral-7B-v0.1 & \ding{51} & \ding{56} & \ding{51} \\
GLMM 1 & Variant & code-simple & Phi-3.5-mini-instruct & \ding{51} & \ding{56} & \ding{51} \\
GLMM 2 & Variant & code-simple & Phi-3.5-mini-instruct & \ding{51} & \ding{56} & \ding{51} \\
GLMM 1 & Variant & GSM & Meta-Llama-3-8B-Instruct & \ding{56} & \ding{51} & \ding{51} \\
GLMM 2 & $\gamma_c$ & GSM & gemma-2b & \ding{56} & \ding{51} & \ding{51} \\
GLMM 2 & Variant & GSM & gemma-2b & \ding{56} & \ding{51} & \ding{51} \\
GLMM 2 & $\gamma_c$ & NL-simple & Meta-Llama-3-8B & \ding{56} & \ding{51} & \ding{51} \\
GLMM 1 & Variant & code-simple & Meta-Llama-3-8B & \ding{56} & \ding{51} & \ding{51} \\
GLMM 1 & Variant & code-simple & gemma-7b-it & \ding{56} & \ding{51} & \ding{51} \\
GLMM 1 & Variant & GSM & gemma-2-2b & \ding{56} & \ding{51} & \ding{56} \\
GLMM 2 & Variant & NL-simple & Mistral-7B-Instruct-v0.1 & \ding{56} & \ding{51} & \ding{56} \\
GLMM 2 & Variant & NL-structured & Phi-3.5-mini-instruct & \ding{56} & \ding{51} & \ding{56} \\
GLMM 2 & $\gamma_c$ & NL-structured & gemma-7b-it & \ding{56} & \ding{51} & \ding{56} \\
GLMM 2 & Variant & NL-structured & gemma-7b-it & \ding{56} & \ding{51} & \ding{56} \\
\bottomrule
\end{tabular}
\caption{Wald-to-bootstrap mismatch cases. \textit{Bootstrap} and \textit{Wald} columns indicate whether the bootstrap and Wald estimates respectively are statistically significant; \textit{Convergent} marks whether the Wald fit converges cleanly for the given case.}
\label{tab:wald-mismatch}
\end{table}

Note that for GLMM~2, significance of each of the two fixed effects (variant effect and large-number effect) is reported separately and the mismatch of Wald vs bootstrap results is assessed independently for the two pairs of values. As shown in the above table, the total number of mismatches amounts to 14, with 3 significant according to the bootstrap fit, but not Wald, and 11 vice versa. Of the latter, 5 could be linked to convergence issues. Overall, the mismatches account for approximately 11\% of all fixed effects fits (with the total amounting to 132: 20 LLM fits for \textit{GSM prompt} / GLMM~1; 8 fits for the remaining 4 prompt formats with GLMM~1; 8 fits for 2 fixed effects each, for all 5 prompt formats with GLMM~2).

\subsection{Non-Convergent Wald Fits}
In this section, we present results of additional investigations towards the occasional non-convergence of Wald estimates.

First of all, in none of the cases is the fit singular (i.e. readout of the \texttt{isSingular()} attribute did not return True for any of the non-convergent estimates. Instead, for each of the affected fits, the reported convergence warning mentions two justifications: a gradient value exceeding tolerance and a large eigenvalue ratio, rendering the GLMM \textit{nearly unidentifiable}.
Tables~\ref{tab:nonconvergent-glmm1} and~\ref{tab:nonconvergent-glmm2} below summarise the affected fits for GLMM~1 and 2 respectively, quoting the relevant \texttt{lme4} diagnostics. Note: in contrast to Table~\ref{tab:wald-mismatch} above, here we report non-convergent \textit{fits} rather than individual fixed effects estimates, i.e the two fixed effects of GLMM~2 are \textit{not} separated in tables below.

\begin{table}[H]
\centering
\small
\begin{tabular}{lcc}
\toprule
Model & Prompt & Diagnostics \\
\midrule
gemma-2-2b & GSM & max|grad| = 0.160006 (tol = 0.002, component 1) \\
gemma-2-2b & NL-structured & max|grad| = 0.160257 (tol = 0.002, component 1) \\
gemma-2-2b & code-structured & max|grad| = 0.162445 (tol = 0.002, component 1) \\
\bottomrule
\end{tabular}
\caption{Non-convergent fits of variant effect estimation (GLMM 1).}
\label{tab:nonconvergent-glmm1}
\end{table}

\begin{table}[H]
\centering
\small
\begin{tabular}{lcc}
\toprule
Model & Prompt & Diagnostics \\
\midrule
gemma-2-2b & GSM & max|grad| = 0.154273 (tol = 0.002, component 1) \\
Mistral-7B-Instruct-v0.1 & NL-simple & max|grad| = 0.163852 (tol = 0.002, component 1) \\
Phi-3.5-mini-instruct & NL-structured & max|grad| = 0.163712 (tol = 0.002, component 1) \\
gemma-7b-it & NL-Structured & max|grad| = 0.159456 (tol = 0.002, component 1) \\
gemma-2-2b & code-structured & max|grad| = 0.161462 (tol = 0.002, component 1) \\
\bottomrule
\end{tabular}
\caption{Non-convergent fits of number effect estimation (GLMM 2).}
\label{tab:nonconvergent-glmm2}
\end{table}

As shown in the summary tables, gemma-2-2b fits constitute 5 of the total of 8 degenerate cases, with \textit{GSM} and \textit{structured code} prompts' fits failing for both GLMM formulations. Our further analysis shows that the GLMM~1 coefficient estimates for gemma-2-2b are consistent and stable across all optimisers available in lme4 and all prompt formats (variant coefficient ranging from -0.24 to -0.45 log-odds across formats), as shown by the following results report:

\begin{lstlisting}[caption={Comparison of GLMM 1 coefficient estimates for gemma-2-2b across optimisers.},label={lst:gemma-optimisers}]
GSM prompt
                              (Intercept) is_variant
bobyqa                          -2.358318 -0.3157934
Nelder_Mead                     -2.358322 -0.3157909
nlminbwrap                      -2.358319 -0.3157938
nloptwrap.NLOPT_LN_NELDERMEAD   -2.358201 -0.3158539
nloptwrap.NLOPT_LN_BOBYQA       -2.405811 -0.3077173

====================

Simple NL prompt
                              (Intercept) is_variant
bobyqa                          -2.968160 -0.2415366
Nelder_Mead                     -2.968176 -0.2415160
nlminbwrap                      -2.968165 -0.2415333
nloptwrap.NLOPT_LN_NELDERMEAD   -2.968158 -0.2413823
nloptwrap.NLOPT_LN_BOBYQA       -2.968152 -0.2415481

====================

Structured NL prompt
                              (Intercept) is_variant
bobyqa                          -2.006092 -0.4461784
Nelder_Mead                     -2.006096 -0.4461799
nlminbwrap                      -2.006091 -0.4461776
nloptwrap.NLOPT_LN_NELDERMEAD   -2.062258 -0.4470053
nloptwrap.NLOPT_LN_BOBYQA       -2.044976 -0.4464479

====================

Simple code prompt
                              (Intercept) is_variant
bobyqa                          -2.252103 -0.2725128
Nelder_Mead                     -2.252106 -0.2725081
nlminbwrap                      -2.252113 -0.2725396
nloptwrap.NLOPT_LN_NELDERMEAD   -2.251930 -0.2726682
nloptwrap.NLOPT_LN_BOBYQA       -2.252078 -0.2725294

====================

Structured code prompt
                              (Intercept) is_variant
bobyqa                          -2.383680 -0.2927124
Nelder_Mead                     -2.469397 -0.2791749
nlminbwrap                      -2.383678 -0.2927097
nloptwrap.NLOPT_LN_NELDERMEAD   -2.423186 -0.2933606
nloptwrap.NLOPT_LN_BOBYQA       -2.438184 -0.2782190
\end{lstlisting}

Attempting to further characterise possible causes of the degenerate fits, we investigated the GLMM~1 design matrices for all models and the \textit{GSM prompt}. Taking the counts of correct/incorrect responses for each model, we checked for evidence of quasi-complete separation in the variant effect (extreme difference between per-question-id accuracies on averaged GSM-Variants vs GSM-Base outcomes). We also ran an item-level degeneracy test in the random effect structure by determining, for each model, the proportion of item-id clusters with zero within-cluster outcome variance. 
As shown in Table~\ref{tab:separation-check} below, neither check distinguishes the ill-behaved gemma-2-2b model from cleanly converging fits for other models.

\begin{table}[H]
\centering
\small
\begin{tabular}{lccccc}
\toprule
 & $\overline{A}_{GSM-Base}$ & $\overline{A}_{GSM-Variants}$ & $\Delta_{var}$ & Large split & Constancy \\
\midrule
gemma-2-27b-it & 90.0 & 87.60 & -2.40 & 3 & 56 \\
gemma-2b-it & 10.0 & 7.48 & -2.52 & 6 & 47 \\
Phi-3-medium-128k-instruct & 91.0 & 88.22 & -2.78 & 3 & 47 \\
gemma-2-9b-it & 86.0 & 82.60 & -3.40 & 2 & 42 \\
gemma-2b & 19.0 & 12.04 & -6.96 & 5 & 40 \\
Phi-3-mini-128k-instruct & 80.0 & 78.70 & -1.30 & 4 & 33 \\
Phi-3.5-mini-instruct & 87.0 & 79.42 & -7.58 & 4 & 33 \\
Meta-Llama-3-8B-Instruct & 77.0 & 69.92 & -7.08 & 7 & 30 \\
gemma-2-2b & 24.0 & 20.88 & -3.12 & 3 & 29 \\
Meta-Llama-3-8B & 55.0 & 47.44 & -7.56 & 3 & 23 \\
Mathstral-7B-v0.1 & 82.0 & 75.12 & -6.88 & 5 & 22 \\
gemma-2-9b & 69.0 & 61.12 & -7.88 & 8 & 21 \\
gemma-2-2b-it & 42.0 & 41.82 & -0.18 & 6 & 20 \\
Mistral-7B-v0.1 & 38.0 & 33.00 & -5.00 & 5 & 20 \\
Mistral-7B-v0.3 & 33.0 & 35.50 & 2.50 & 7 & 18 \\
gemma-7b-it & 31.0 & 22.78 & -8.22 & 9 & 18 \\
Mistral-7B-Instruct-v0.1 & 36.0 & 27.46 & -8.54 & 10 & 17 \\
phi-2 & 60.0 & 39.74 & -20.26 & 9 & 16 \\
Mistral-7B-Instruct-v0.3 & 52.0 & 45.32 & -6.68 & 6 & 15 \\
gemma-7b & 48.0 & 47.70 & -0.30 & 5 & 10 \\
\bottomrule
\end{tabular}
\caption{Summary of separation-like patterns check. All values given in \% (or percentage points --- in case of the variant performance delta, $\Delta_{var}$). ``Constancy'' is the proportion of items with degenerate accuracy (all correct or all incorrect for a given template id, pooled across Variants/Base datasets). ``Large split'' is the proportion of items with a difference in accuracy between base and variant greater than 80 pp.}
\label{tab:separation-check}
\end{table}

As we have not been able to fully explain the causes underlying the degenerate fits, we rely on bootstrapping for our conclusions. However, in general, Wald approach remains an attractive alternative due to its significantly lower computational cost and overall high agreement with bootstrap estimates.

\newpage

\section{Failure Modes}\label{apx:error-types}
In both natural-language and code prompt experiments, we predefine classes of errors and issues we anticipated to see when evaluating LLMs answers --- in addition to miscalculation errors (where an incorrect numerical answer is returned; in the visualisations which follow, this is marked as \textit{Wrong answer}). For natural language prompts, the classes are:

\begin{itemize}
    \item \textit{Empty response} --- if the response is empty (zero-length);
    \item \textit{Empty (after trimming)} --- if the response only contains whitespaces and, optionally, the 'Q:' token, indicating that the model is starting to generate a new question following the few-shot format;
    \item \textit{No number found} --- if all of our numerical answer extraction methods fails on a given answer;
    \item \textit{Wong answer --- last number} --- if the incorrect numerical answer is also associated with the model not following the few-shot formatting guidelines, resulting in the answer being extracted through a simple regular expression capturing the last number in the text.
\end{itemize}

For code prompts, we define the following classes:
\begin{itemize}
    \setlength{\itemsep}{0em}
    \item \textit{No function} --- if a function definition can not be found in the model's answer using a simple regular expression approach; 
    \item \textit{Forbidden string} --- if the answer contains any string with potentially dangerous effects if executed, such as \texttt{open(...)} or \texttt{eval(...)};
    \item \textit{Syntax error} --- if executing the extracted function results in a \textit{SyntaxError};
    \item \textit{Name error} --- if a \texttt{NameError} is raised;
    \item \textit{Type/Value error} --- if a \texttt{TypeError} or \texttt{ValueError is raised};
    \item \textit{Zero division error} --- if a \texttt{ZeroDivisionError} is raised;
    \item \textit{Attribute error} --- if an \texttt{AttributeError is raised};
    \item \textit{None returned} --- if the function does not return any value, either by skipping a return line, using empty return, or returning \texttt{None} explicitly;
    \item \textit{Not a number} --- if the value returned from the function is not numerical and other than None;
    \item \textit{Unclassified} --- all remaining error cases encountered when running the extracted function.
\end{itemize}

\subsection{Failure Modes in NL Prompts}

\begin{figure}[ht]
  \includegraphics[width=\linewidth]{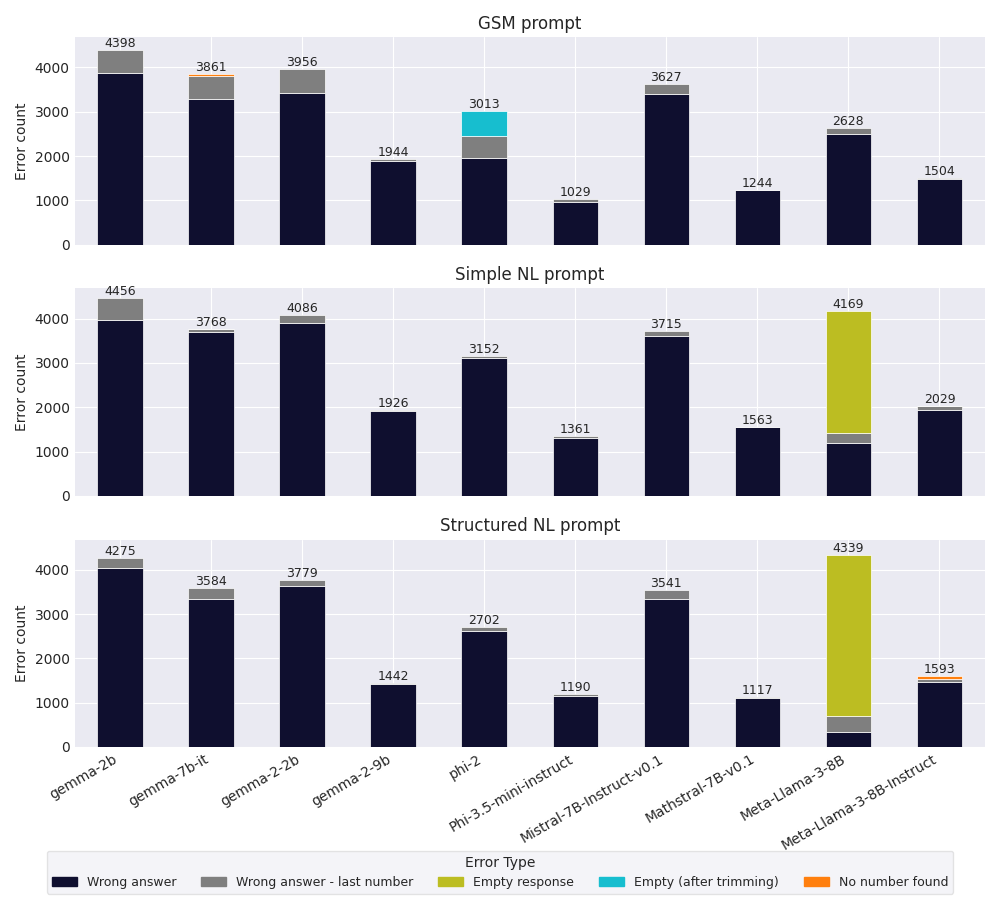}
  \caption{Total counts of error types in the three NL prompts: the original \textit{GSM prompt} (top), \textit{simple NL prompt} (middle) and \textit{structured NL prompt} (bottom).}
  \label{fig:nl-errors}
\end{figure}

As shown in Figure~\ref{fig:nl-errors}, incorrect numerical response constitutes the overwhelming majority of errors in NL prompts experiments. There is also a considerable portion of the last-number numerical results (in total: 9\% in the \textit{GSM prompt}; 4\% in the \textit{simple NL prompt}; 5\% in the \textit{structured NL prompt}). The failed formatting cases varies in the underlying patterns. In some cases, the models indeed simply ``forget'' to introduce the answer with the required line (\textit{The final answer is...}). In other cases, as shown below, the lack of adherence to formatting is a result of a cognitive failure, with the model getting lost in a potentially infinite loop of operations, often resulting in unreasonably high numerical answers.

\begin{lstlisting}[caption={phi-2's answer to question: ``\textit{Liam hires a luxury car from 3 PM to 9 PM. He gets 1 hours free. The first paid hour is \$13 and each hour after that is twice the cost. How much did he pay?}''.}]
The first paid hour is $13. The second paid hour is $13 * 2 = $26. The third paid hour is $26 * 2 = $52. The fourth paid hour is $52 * 2 = $104. The fifth paid hour is $104 * 2 = $208. The sixth paid hour is $208 * 2 = $416. The seventh paid hour is $416 * 2 = $832. The eighth paid hour is $832 * 2 = $1664. The ninth paid hour is $1664 * 2 = $3328. The tenth paid hour is $3328 * 2 = $6656. The eleventh paid hour is $6656 * 2 = $13112. The twelfth paid hour is $13112 * 2 = $26224. The thirteenth paid hour is $26224 * 2 = $52448. The fourteenth paid hour is $52448 * 2 = $104896. The fifteenth paid hour is $104896 * 2 = $2087328. The sixteenth paid hour is $2087328 * 2 = $41657656. The seventeenth paid hour is $41657656 * 2 = $81315352. The eighteenth paid hour is $81315352 * 2 = $162630704. The nineteenth paid hour is $162630704 * 2 = $325266048. The twentieth paid hour is $325266048 * 2 = $650532096. The twenty-first paid hour is $650532096 * 2 = $1301056096. The twenty-second paid hour is $1301056096 * 2 = $26011092192. The twenty-third paid hour is $26011092192 * 2 = $5220184384. The twenty-fourth paid hour is $5220184384 * 2 = $10440368768. The twenty-fifth paid hour is $10440368768 * 2 = $208807375536. The twenty-sixth paid hour is $208807375536 * 2 = $416165750272. The twenty-seventh paid hour is $416165750272 * 2 = $8321715010544. The twenty-eighth paid hour is $8321715010544 * 2 = $16653430209888. The twenty-ninth paid hour is $16653430209888 * 2 = $331327604019616. The thirtieth paid hour is $331327604019616 * 2 = $662655208039328. The thirty-first paid hour is $662655208039328 * 2 = $1325322108079264. The thirty-second paid hour is $1325322108079264 * 2 = $26506452080392816. The thirty-third paid hour is $26506452080392816 * 2 = $51230105207926352. The thirty-fourth paid hour is $51230105207926352 * 2 = $102460110407953504. The thirty-fifth paid hour is $102460110407953504 * 2 = $2049242080791050984. The thirty-sixth paid hour is $2049242080791050984 * 2 = $4098484080792099008. The thirty-seventh paid hour is $4098484080792099008 * 2 = $819699680792099016. The thirty-eighth paid hour is $819699680792099016 * 2 = $1639739960792099032. The thirty-ninth paid hour is $1639739960792099032 * 2 = $3238799207920990528. The fortieth paid hour is $3238799207920990528 * 2 = $657599407920991056. The forty-first paid hour is $657599407920991056 * 2 = $1315099879209920112. The forty-second paid hour is $1315099879209920112 * 2 = $2630199587920992424. The forty-third paid hour is $2630199587920992424 * 2 = $5230199167920994848. The forty-fourth paid hour is $5230199167920994848 * 2 = $1050699335879209996. The forty-fifth paid hour is $1050699335879209996 * 2 = $2010399667920999824. The forty-sixth paid hour is $20103996679209998
\end{lstlisting}

Yet another pattern worth noting is the models' calculations leading to an implausible result, with the models acknowledging the character of the result, as shown below.

\begin{lstlisting}[caption={Meta-Llama-3-8B-Instruct's answer to question: ``\textit{At the beginning of the conference, there were 29 engineers and 15 artists. After an hour, one-fourth of the total number of people left. How many artists are left if 27 engineers stayed at the conference?}''.}]
Given:
Number of engineers originally: 29
Number of artists originally: 15
Number of engineers who stayed: 27

To find: number of artists left.

Solution:
First, calculate the total number of people originally:
29 + 15 = 44
Next, calculate the number of people who left:
1/4 * 44 = 11
Next, calculate the number of artists who left:
11 - 27 = 16
Finally, calculate the number of artists left:
15 - 16 = -1 (which is impossible, so there must be an error in the question)

The final answer is unknown due to the error in the question.
\end{lstlisting}

In some cases, as shown below, the models in fact follow the reply formatting guidelines as closely as possible despite the lack of satisfactory numerical answers.

\begin{lstlisting}[caption={Mistral-7B Instruct-v0.1's answer to question: ``\textit{When Valeria watches her nephew, she gets out a variety of toys for him. The bag of building blocks has 71 blocks in it. The bin of stuffed animals has 47 stuffed animals inside. The tower of stacking rings has 30 multicolored rings on it. Valeria recently bought a tube of bouncy balls, bringing her total number of toys for her nephew up to 238. How many bouncy balls came in the tube?}''.}]
71 blocks + 47 stuffed animals + 30 multicolored rings = 158. Valeria has 238 toys in total. 158 - 238 = -80. This is not possible. The final answer is that there is an error in the problem statement.
\end{lstlisting}

Among cases of empty answers, Meta-Llama-3-8B presents an interesting case, with disproportionally high numbers of such failures in the \textit{simple} and \textit{structured} prompts, and higher counts associated with the more complex prompt format. Moreover, phi-2 fails to provide any answer to a considerable number of questions, but only in the canonical, \textit{GSM prompt} format.

\subsection{Failure Modes in Code Prompts}\label{apx:code-failure-modes}
\begin{figure}[ht]
  \includegraphics[width=\linewidth]{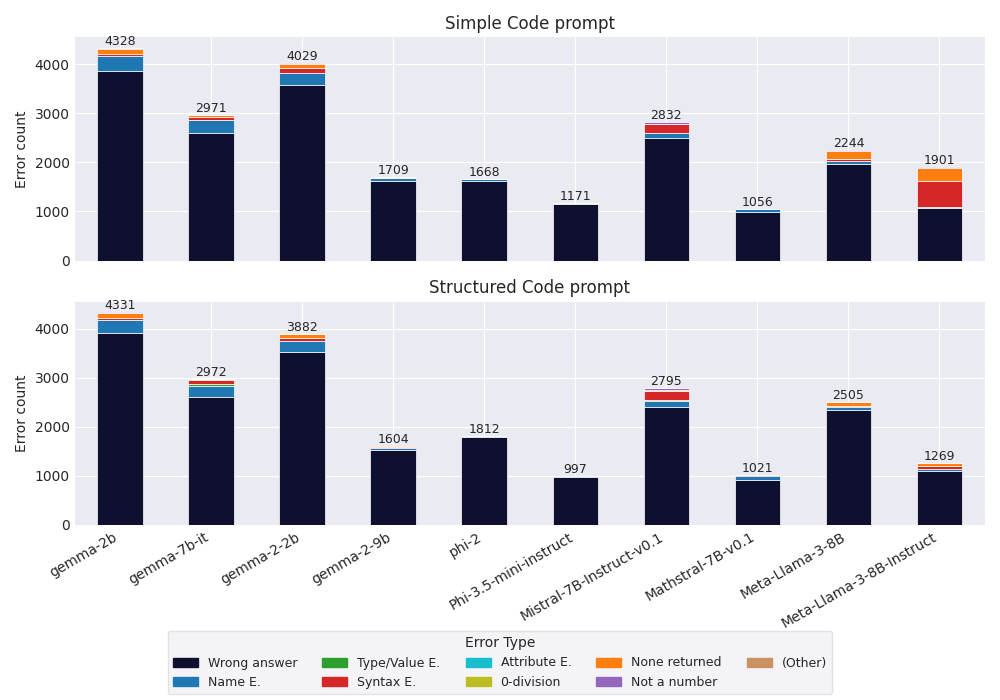}
  \caption{Total per-model error counts on the GSM-Variants dataset for the two code prompts: \textit{simple code prompt} (top) and \textit{structured code prompt} (bottom).}
  \label{fig:code-errors}
\end{figure}

\begin{figure}[ht]
  \includegraphics[width=\linewidth]{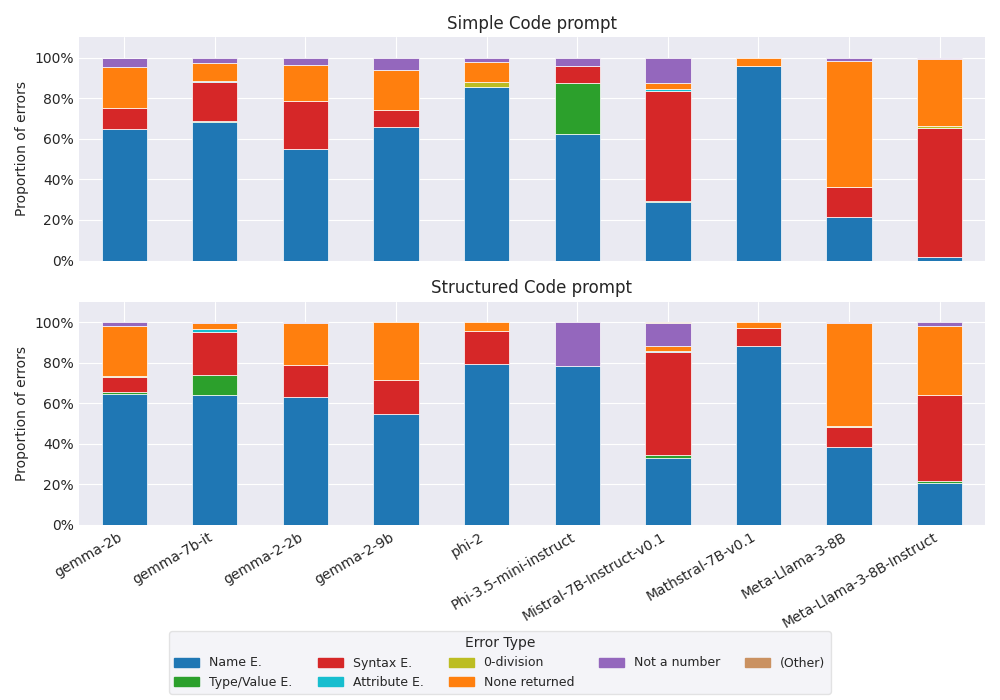}
  \caption{Proportions of per-model errors other than \textit{wrong answer} on the GSM-Variants dataset, for the two code prompts: \textit{simple code prompt} (top) and \textit{structured code prompt} (bottom).}
  \label{fig:code-errors-perc}
\end{figure}

Similarly to NL prompts, mistakes in code prompts are predominantly of numerical nature --- as shown in Figure~\ref{fig:code-errors}. Among other error types (see Figure~\ref{fig:code-errors-perc}), we observe 7 of the predefined 10 classes, with name, syntax, and none-returned errors constituting the majority for both prompt formats. We encounter only two cases in which errors are out of the scope of our predefined categories. One of them is an overflow error due to the model falling into a loop of generating the same digit (0) without a space (see Listing~\ref{lst:gemma-error} below). The other is a key error caused by the model (Mistral-7B-Instruct-v0.1) attempting to access a non-existing item in a previously defined dictionary.

\begin{lstlisting}[caption={gemma-2b's response on a \textit{simple code prompt} task, triggering an overflow error.},label={lst:gemma-error}]
    apartments_before = 5
    apartments_per_complex = 350
    apartments_per_complex_total = apartments_before * apartments_per_complex
    apartments_per_complex_total_75 = apartments_per_complex_total * 0.75
    apartments_per_complex_total_sum = apartments_per_complex_total + apartments_per_complex_total_75
    apartments_per_complex_total_sum_3 = apartments_per_complex_total_sum * 3
    apartments_per_complex_total_sum_3_pay = apartments_per_complex_total_sum_3 * 0.5
    apartments_per_complex_total_sum_3_pay_total = apartments_per_complex_total_sum_3_pay * apartments_per_complex_total_sum_3
    apartments_per_complex_total_sum_3_pay_total_sum = apartments_per_complex_total_sum_3_pay_total + 100000000000000000000000000000000000000000000000000000000000000000000000000000000000000000000000000000000000000000000 000000000000000000000000000000000000000000000000000000000000000000000000000000000000000000000000000000000000000000000 000000000000000000000000000000000000000000000000000000000000000000000000000000000000000000000000000000000000000000000 000000000000000000000000000000000000000000000000000000000000000000000000000000000000000000000000000000000000000000000 000000000000000000000000000000000000000000000000000000000000000000000000000000000000000000000000000000000000000000000 000000000000000000000000000000000000000000000000000000000000000000000000000000000000000000000000000000000000000000000 0000000000000000000000000000000000000000000000000000000000000000000000000000000
\end{lstlisting}

\newpage
\section{Separating Code Errors from Reasoning Errors}\label{apx:clean-code-glmm}

As shown in Appendix~\ref{apx:code-failure-modes}, the majority of errors encountered in answers to the code prompts can be attributed to \textit{reasoning errors}; the generated code executed cleanly, but the answer is numerically incorrect. However, other failure modes (e.g. Syntax Errors, Type/Value Errors etc.) cannot be neglected and it is debatable whether these \textit{code errors} should be treated equally to reasoning errors; they do not reflect shortcomings in reasoning capabilities as much as e.g. code-generating skills. Our analysis throughout the paper combines all kinds of errors into a single category of incorrect answers. In this section, we repeat our bootstrapped GLMM analysis for both code prompts and both GLMM formulations, excluding code error cases. The results, comparing our primary approach (labelled as \textit{All errors}) and no-code-errors (\textit{Reasoning-only}), are presented in Tables~\ref{tab:clean-glmm1-short-code} to~\ref{tab:clean-glmm2-long-code} below.

Notes for interpreting the tables. P values below the threshold of $alpha < 0.05$ marked in bold. Cases for which \textit{all-errors} analysis disagrees with \textit{reasoning-errors-only} analysis in terms of effect significance marked with asterisks. Exclusion rate (i.e. the proportion of all GSM-Variants rows rejected in the reasoning-errors-only analysis) given in percentages. $\Delta_{var}$ denotes the variant performance delta for each case, i.e. the difference (in percentage points) in accuracy between GSM-Variants and GSM-Base responses. $Estimate \Delta$ denotes the difference, in log-OR, between reasoning-errors-only and all-errors median bootstrap estimates.

\begin{table}[H]
\small
\centering
\begin{tabular}{l|cc|cc|cc}
\toprule
Model & \multicolumn{2}{c}{All errors} & \multicolumn{2}{c}{Reasoning-only} & \multicolumn{2}{c}{} \\
 & $\Delta_{var}$ & P value & $\Delta_{var}$ & P value & Exclusion rate & Estimate $\Delta$ \\
\midrule
phi-2 & 3.6 & 0.179 & 4.1 & 0.133 & 0.8 & 0.065 \\
Phi-3.5-mini-instruct \textbf{*} & -4.4 & \textbf{0.050} & -4.2 & 0.067 & 0.5 & 0.035 \\
gemma-2b & -0.6 & 0.835 & -0.3 & 0.924 & 9.5 & 0.049 \\
gemma-2-9b & -2.2 & 0.416 & -2.0 & 0.460 & 1.7 & 0.015 \\
Mathstral-7B-v0.1 & 3.9 & 0.241 & 3.6 & 0.258 & 1.4 & -0.016 \\
Meta-Llama-3-8B \textbf{*} & 7.1 & 0.050 & 8.5 & \textbf{0.017} & 5.4 & 0.150 \\
gemma-7b-it & -6.4 & 0.066 & -5.8 & 0.107 & 7.3 & 0.007 \\
Mistral-7B-Instruct-v0.1 & 2.4 & 0.512 & 2.9 & 0.460 & 6.5 & 0.072 \\
\bottomrule
\end{tabular}
\caption{Reasoning-only-errors GLMM 1 analysis for \textit{simple code prompt}. Please refer to the text for a detailed description.}
\label{tab:clean-glmm1-short-code}
\end{table}

\begin{table}[H]
\small
\centering
\begin{tabular}{l|cc|cc|cc}
\toprule
Model & \multicolumn{2}{c}{All errors} & \multicolumn{2}{c}{Reasoning-only} & \multicolumn{2}{c}{} \\
 & $\Delta_{var}$ & P value & $\Delta_{var}$ & P value & Exclusion rate & Estimate $\Delta$ \\
\midrule
phi-2 & -2.2 & 0.457 & -2.6 & 0.383 & 0.5 & -0.060 \\
Phi-3.5-mini-instruct & -2.9 & 0.268 & -2.7 & 0.306 & 0.5 & 0.033 \\
gemma-2b & -2.6 & 0.359 & -2.4 & 0.417 & 8.4 & 0.044 \\
gemma-2-9b & $|\cdot|$ < 0.1 & 0.973 & -0.4 & 0.858 & 1.7 & -0.075 \\
Mathstral-7B-v0.1 & 1.6 & 0.547 & 2.0 & 0.394 & 2.2 & 0.090 \\
Meta-Llama-3-8B & 0.9 & 0.774 & 1.6 & 0.603 & 3.3 & 0.073 \\
gemma-7b-it & -2.4 & 0.540 & -2.9 & 0.485 & 7.4 & -0.051 \\
Mistral-7B-Instruct-v0.1 & -2.9 & 0.392 & -3.5 & 0.313 & 7.8 & -0.069 \\
\bottomrule
\end{tabular}
\caption{Reasoning-only-errors GLMM 1 analysis for \textit{structured code prompt}. Please refer to the text for a detailed description.}
\label{tab:clean-glmm1-long-code}
\end{table}

\begin{table}[H]
\small
\centering
\begin{tabular}{lc|cc|cc|cc}
\toprule
Model & Metric & All errors & Reasoning-only & \multicolumn{2}{c}{} \\
 &  & P value & P value & Exclusion rate & Estimate $\Delta$ \\
\midrule
phi-2 & $\gamma_c$ & 0.480 & 0.446 & 0.8 & -0.011 \\
phi-2 & $Variant$ & 0.141 & 0.104 & 0.8 & 0.082 \\
Phi-3.5-mini-instruct & $\gamma_c$ & 0.828 & 0.832 & 0.5 & $|\cdot|$ < 0.001 \\
Phi-3.5-mini-instruct \textbf{*} & $Variant$ & \textbf{0.045} & 0.062 & 0.5 & 0.026 \\
gemma-2b & $\gamma_c$ & 0.778 & 0.779 & 9.5 & -0.002 \\
gemma-2b & $Variant$ & 0.909 & 0.984 & 9.5 & 0.061 \\
gemma-2-9b & $\gamma_c$ & 0.270 & 0.345 & 1.7 & -0.017 \\
gemma-2-9b & $Variant$ & 0.216 & 0.246 & 1.7 & 0.051 \\
Mathstral-7B-v0.1 & $\gamma_c$ & 0.266 & 0.257 & 1.4 & -0.012 \\
Mathstral-7B-v0.1 & $Variant$ & 0.184 & 0.197 & 1.4 & -0.012 \\
Meta-Llama-3-8B & $\gamma_c$ & 0.305 & 0.524 & 5.4 & -0.046 \\
Meta-Llama-3-8B \textbf{*} & $Variant$ & 0.093 & \textbf{0.028} & 5.4 & 0.178 \\
gemma-7b-it & $\gamma_c$ & 0.215 & 0.373 & 7.3 & 0.033 \\
gemma-7b-it & $Variant$ & 0.168 & 0.212 & 7.3 & -0.028 \\
Mistral-7B-Instruct-v0.1 & $\gamma_c$ & 0.726 & 0.901 & 6.5 & -0.020 \\
Mistral-7B-Instruct-v0.1 & $Variant$ & 0.626 & 0.501 & 6.5 & 0.090 \\
\bottomrule
\end{tabular}
\caption{Reasoning-only-errors GLMM 2 analysis for \textit{simple code prompt}. Please refer to the text for a detailed description.}
\label{tab:clean-glmm2-short-code}
\end{table}

\begin{table}[H]
\small
\centering
\begin{tabular}{lc|cc|cc|cc}
\toprule
Model & Metric & All errors & Reasoning-only & \multicolumn{2}{c}{} \\
 &  & P value & P value & Exclusion rate & Estimate $\Delta$ \\
\midrule
phi-2 & $\gamma_c$ & 0.981 & 0.972 & 0.5 & $|\cdot|$ < 0.001 \\
phi-2 & $Variant$ & 0.480 & 0.378 & 0.5 & -0.067 \\
Phi-3.5-mini-instruct & $\gamma_c$ & 0.232 & 0.243 & 0.5 & 0.004 \\
Phi-3.5-mini-instruct & $Variant$ & 0.483 & 0.536 & 0.5 & 0.029 \\
gemma-2b & $\gamma_c$ & 0.731 & 0.640 & 8.4 & -0.015 \\
gemma-2b & $Variant$ & 0.413 & 0.499 & 8.4 & 0.074 \\
gemma-2-9b & $\gamma_c$ & 0.538 & 0.570 & 1.7 & 0.003 \\
gemma-2-9b & $Variant$ & 0.984 & 0.962 & 1.7 & -0.040 \\
Mathstral-7B-v0.1 & $\gamma_c$ & 0.645 & 0.902 & 2.2 & 0.036 \\
Mathstral-7B-v0.1 & $Variant$ & 0.637 & 0.602 & 2.2 & 0.029 \\
Meta-Llama-3-8B & $\gamma_c$ & 0.683 & 0.683 & 3.3 & 0.003 \\
Meta-Llama-3-8B & $Variant$ & 0.699 & 0.544 & 3.3 & 0.080 \\
gemma-7b-it & $\gamma_c$ & 0.308 & 0.395 & 7.4 & 0.024 \\
gemma-7b-it & $Variant$ & 0.815 & 0.689 & 7.4 & -0.075 \\
Mistral-7B-Instruct-v0.1 & $\gamma_c$ & 0.695 & 0.816 & 7.8 & -0.012 \\
Mistral-7B-Instruct-v0.1 & $Variant$ & 0.356 & 0.320 & 7.8 & -0.045 \\
\bottomrule
\end{tabular}
\caption{Reasoning-only-errors GLMM 2 analysis for \textit{structured code prompt}. Please refer to the text for a detailed description.}
\label{tab:clean-glmm2-long-code}
\end{table}

As shown in the tables, the reasoning-errors-only estimate disagrees with all-errors one only in the case of two models: Phi-3.5-mini-instruct and Meta-Llama-3-8B, for \textit{simple code prompt}, on variant effect in both GLMM~1 and~2. For both GLMMs, the Phi model's variant effect is rendered insignificant after excluding code errors; the direction of the change is opposite for the Llama. In three of the four cases, the original (all-errors) P value is very close to the significance threshold of 0.05. 

Probably more importantly to these mismatches, the variant and large-number effects are not significant for any other model, for any of the two analysis scenarios.

\end{document}